\documentclass[11pt]{article}

\usepackage[final]{latex/acl}

\usepackage{times}
\usepackage{latexsym}

\usepackage[T1]{fontenc}

\usepackage[utf8]{inputenc}

\usepackage{microtype}

\usepackage{inconsolata}

\usepackage{graphicx}

\usepackage{adjustbox}
\usepackage{ amssymb }
\usepackage{listings}
\usepackage{pgfplots}
\usepackage{subcaption}
\pgfplotsset{compat=1.18}
 
\usepackage{booktabs}

\usepackage[dvipsnames]{xcolor}

\newcommand{\localbest}[1]{\mathbf{#1}}

\usepackage{array}

\usepackage{multirow}

\usepackage{amsmath}

\usepackage{xcolor}
\usepackage[table]{xcolor}
\usepackage[dvipsnames]{xcolor}
\usepackage{tabularx}

\title{Breaking Bureaucracy: Evaluating open-source LLMs for legal document review
}

\author{Farrukh Baratov \\
  Leiden University\\ Leiden, the Netherlands\\
  \small{\texttt{f.s.baratov@liacs.leidenuniv.nl}} \\\And
  Niki van Stein\\
  Leiden University\\ Leiden, the Netherlands\\
  \small{\texttt{n.van.stein@liacs.leidenuniv.nl}} \\\And
  Suzan Verberne\\
  Leiden University\\ Leiden, the Netherlands\\
  \small{\texttt{s.verberne@liacs.leidenuniv.nl}}}

\begin{document}
\maketitle

\begin{abstract}
In this paper, we evaluate open-source generative LLMs on legal Natural Language Inference (NLI). 
Legal inspectorial processes take place in specific domains and often deal with confidential data. This creates a need for working with local models that do not require labeled training data.
We evaluate our models on the ContractNLI benchmark and two NLI4Wills datasets. We successfully reproduce the baseline for the task (Span NLI BERT) and we evaluate multiple open-source LLMs on the same task. We analyze the invalid rate of the models, and their stability across temperature settings and domains. 
Among the generative models, Gemma-4 26B performs the best, reaching an accuracy of 81.2\%, even outperforming the supervised model on one metric. On accuracy, it is not possible to beat the supervised model with zero-shot approaches. Qwen-3.6 35B performs well on both ContractNLI and additional datasets in the legal wills domain. 
Our findings indicate that zero-shot, open-source, generative LLMs are a viable alternative for real-world legal NLI when no supervised data is available. 
{Our code is available at \url{https://github.com/fbaratov/contractnli-llms}.}
\end{abstract}

\section{Introduction}
Document review is a task that 
involves human experts investing considerable hours reading and reviewing a number of legal documents, and is a large burden on legal processes for organizations \cite{nithya2024ai}. This includes government inspectorates, whose administrative inspections are heavily focused on document review for regulatory non-compliance detection.
As a result,
there has been an interest from inspectorates in applying computational methods to their inspections, hence reducing the workload for human experts.

One potential method for computer-assisted document review is natural language inference (NLI). 
NLI is especially fitting to assist inspectors in their responsibility of reviewing documents with regards to government regulations in order to assess their regulatory compliance. 
However, there are several obstacles in effectively integrating deep learning-based methods into inspectorial processes. First, inspections are expected to be confidential, meaning that API-based proprietary models would pose a security risk. 
Second, inspectorial processes take place in specific legal domains, such as education and transport, thus supervised methods require substantial data labeling for each newly presented domain. This issue is often exacerbated by the bureaucratic obstacles of gaining access to potentially viable data,
eliminating the option of extensive training or fine-tuning supervised methods.


In this paper, we evaluate open-source LLMs on ContractNLI \cite{koreeda2021contractnli}, an English-language legal NLI dataset, to investigate the potential of these models in settings without labelled training data. Open-source LLMs are especially suited for our use case, as they (a) can be run locally, and can thus be trusted with sensitive documents, (b) can be used without extensive fine-tuning through zero-shot and few-shot learning, and (c) 
have accessible inner workings, allowing them to convey predictive uncertainty.

Our contributions are as follows: (1) We reproduce \citet{koreeda2021contractnli}'s model for ContractNLI, Span NLI BERT, in order to verify their results and confirm it as a reliable baseline;
(2) We evaluate several state-of-the-art open-weight quantized LLMs on ContractNLI, finding that they do not meet the supervised results but reach reasonable accuracy (up to 81\%) without any finetuning. We additionally assess generalizability to different legal domains by validating on the NLI4Wills dataset; (3) We conduct a temperature and logit-based stability analyses, finding that the models are robust against changes in temperature, but that logits for most models do not reliably reflect uncertainty. 


\section{Related work}
\paragraph{NLI for legal document review}
\citet{chalesthori2024evaluating} conclude that, based on criteria such as task difficulty, as well as real-world risks and applications, ContractNLI is a good dataset for evaluating AI-assisted decision making. Multiple previous studies, such as \citet{schuster2022stretching}, and \citet{tang2023natural}, have evaluated models on the ContractNLI dataset. \citet{narendra2024enhancing} present results based on GPT4 and a fine-tuned Mixtral implementation. However, they exclude the ``NotMentioned'' labels from their evaluation, as the focus of their study was on ContractNLI's evidence identification task. 
Most recently, \citet{liu2025contracteval} evaluated multiple proprietary and open-source LLMs on the CUAD dataset \cite{cuad}, which focuses on identifying clauses relevant to a specific label. They find that, on this task, proprietary models generally outperform open-source models. They also find that results are improved by larger model size and reasoning capabilities, and that quantization negatively impacts performance. Their results indicate that open-source LLMs need to be fine-tuned to reach the performance of proprietary models. We put open-source zero-shot models to the test for a different task, legal NLI, to investigate their potential in domain-specific settings without labelled data for fine-tuning.

\paragraph{Uncertainty estimation for LLMs}
\citet{shorinwa2025survey} present an overview of current uncertainty quantification (UQ) methods, including white-box methods which assume that the model's architecture is at least partially visible. 
Methods such as predictive entropy, proposed by \citet{shannon1948mathematical} and applied to deep learning by authors such as  \citet{gal2016uncertainty} and \citet{ashukha2020pitfalls}, also consider the output distribution of a given token by assessing the log-probabilities of every possible token. 

\section{Methods}
\subsection{Generative legal NLI}

\paragraph{Output structures} In order to use generative LLMs for the NLI task, we apply the JSON output structures (see Appendix \ref{app:structures}) during inference to improve inference quality and minimize invalid answer rates. The structures consist of a free-text ``explanation'' field, followed by a ``classification'' field that outputs one of the dataset labels. We choose this order so that the explanation will be in the model's context window when it generates a label. This way, the LLMs initially generate a free-text explanation that can be read and interpreted by a human user, and then summarize the outcomes of this explanation into a single label.

\paragraph{Prompts} We evaluate on two prompts, provided in Appendix \ref{app:prompts}, that closely resemble those used by \citet{narendra2024enhancing}. These prompts are relatively simple and allow for the face-value evaluation of LLM performance, focusing more on their inherent capability to handle the challenge posed by our datasets, and less on how performance can be increased through prompt engineering.

\subsection{Uncertainty quantification}

We use two of the token-level white-box UQ methods compiled by \citet{shorinwa2025survey}: Average negative log-probability and maximum negative token log-probability\footnote{
\citet{shorinwa2025survey} refer to these metrics as ``average log-probability'' and ``maximum token log-probability'' respectively. The equations themselves use negative log-probability. Therefore, we refer to them differently for added clarity.
\label{foot:shorinwa}} (Appendix \ref{app:uq-equations} for equations).

\section{Experiments}

\subsection{Datasets}
\paragraph{ContractNLI} We focus on ContractNLI\footnote{\url{https://stanfordnlp.github.io/contract-nli/}} \cite{koreeda2021contractnli} to evaluate our models. ContractNLI is a document-level NLI dataset which defines legal contracts as premises, and a set of 17 different statements as hypotheses.
The dataset contains 607 contracts to be used as premises for the NLI task, with each contract containing annotations for 17 hypotheses.  
The annotations include an NLI label for whether the document entails, contradicts, or is unrelated to the hypothesis. 
We use the test set of 123 documents for our experiments. It contains a total of 2091 document-hypothesis samples, 1188 of which are labelled either ``Contradiction'' or ``Entailment'', and 903 are labelled ``NotMentioned''.
\paragraph{NLI4Wills} The NLI4Wills datasets\footnote{\url{https://github.com/ml4ai/nli4wills-corpus/tree/main/data}} \cite{kwak-et-al-nllp2023-error-analysis} are legal NLI datasets that focus on evaluating the validity of legal will statements in the US states of Idaho and Tennessee. Three inputs are provided: a statement from a will, a condition, and a law. The task is to evaluate whether the law and condition \textit{support}, \textit{refute}, or are \textit{unrelated} to the given statement. NLI4Wills consists of two datasets: NLI4Wills-Idaho and NLI4Wills-Tennessee. NLI4Wills-Idaho contains 154 input samples, and NLI4Wills-Tennessee contains 255 input samples.

\subsection{Baseline reproduction}
To reproduce the Span NLI BERT baseline introduced by \citet{koreeda2021contractnli}, we utilize the code and instructions provided by the authors on GitHub.\footnote{\url{https://github.com/stanfordnlp/contract-nli-bert}} 
We were able to train the model provided by the authors with no significant problems and with minimal changes to the code itself.


\subsection{LLM evaluation}
\paragraph{Models} We evaluate models of varying sizes from four different language families: Gemma-4 (E2B, E4B, 26B) \cite{gemma4}, Ministral-3 (3B, 8B, 14B) \cite{ministral-3}, Qwen-3.5 (4B, 9B, 27B) \cite{qwen3.5}, and Qwen-3.6 (35B) \cite{qwen3.6}. These model families are primarily chosen for their recency and variance in size, which will allow us to compare models from the same family or models of similar sizes. We run all models larger than 20 billion parameters in reasoning mode, as we expect that their larger size will allow them to make better use of reasoning mode when quantized compared to smaller models.

We set the temperature to 0.3 for all models to encourage differences in explanations across seeds, while still keeping stochasticity low enough to keep classifications reliable. We provide output structures for all models except Gemma-4 26B to follow (see Appendix \ref{app:structures}).
We set context windows to 16,384 tokens, and limit the outputs to 5,000 tokens.

We use Ollama to conduct our experiments. All models use 4-bit quantization. As a result, all models can comfortably run on relatively inexpensive consumer-grade hardware.

\paragraph{Evaluation}
We use the three NLI metrics proposed by \citet{koreeda2021contractnli}: the contradiction F1-score F1(C), entailment F1-score F1(E), and overall model accuracy. We micro-average these scores over documents and then macro-average them over labels. We evaluate our results against Span NLI BERT for the all-label setting. Because generative models sometimes lead to invalid output, we modify the evaluation script provided by \citet{koreeda2021contractnli} to handle invalid LLM outputs
by treating them as incorrect. For all datasets, we report the mean and standard deviations of metrics over three seeds, as well as invalid output rates. 


\subsection{Additional analyses}

\textbf{Temperature analysis} We repeat our evaluation for each model on several temperatures between 0.1 and 0.5. Comparing metrics between various temperature settings will help us assess the models' robustness against changes in temperature. \\
\textbf{Uncertainty analysis} For each model-label-seed combination, we perform a Mann-Whitney U Test \cite{mann1947test} to determine whether the distribution of uncertainty metrics for correct answers is stochastically less than the distribution of uncertainty metrics for incorrect answers. 
We use Benjamini-Hochberg false discovery control \cite{benjamini1995controlling} and calculate effect sizes using Rank Biserial Correlation \cite{wendt1972dealing}.
We evaluate primarily only on the logits of the label. We also provide results on (a) the logits of the whole answer and (b) the logit of the label's first token (Appendix \ref{app:logit-analysis-results}). \\
\textbf{NLI4Wills validation} To assess generalizability to other legal domains, we evaluate the best performing configurations for each model on the NLI4Wills datasets. Following \citet{kwak-et-al-nllp2023-error-analysis}, we report results on Precision, Recall, F1-scores, and Accuracy.

\textbf{Ablation study} We perform an ablation study for reasoning and structured outputs on all reasoning-capable models (Appendix \ref{app:model-specific-metrics}).


\begin{table*}[t]
    \centering
    \begin{tabular}{l||ccr|c|c}
        \textbf{Metric} & \textbf{K\&M} & \textbf{Reproduction} & \textbf{Re-Or} & \textbf{O.SD} & \textbf{R.SD} \\
        \hline
        \hline
        mAP      & $0.922\pm0.006$ & $0.919\pm0.005$ & $-0.003$ 
                 & \checkmark & \checkmark \\
        P@R80   & $0.793\pm0.018$ & $0.801\pm0.022$ & $+0.008$ 
                 & \checkmark & \checkmark  \\
        \hline
        Acc & $0.875\pm0.006$ & $0.869\pm0.006$ & $-0.006$
                 & \checkmark & \checkmark \\
        F1 (C)   & $0.357\pm0.039$ & $0.341\pm0.043$ & $-0.034$
                 & \checkmark & \checkmark \\
        F1 (E)   & $0.834\pm0.002$ & $0.826\pm0.004$ & $-0.008$
                 & \texttimes & \texttimes \\
    \end{tabular}
    \caption{\small Reproduction results for Span NLI BERT. ``Re-Or'' indicates the reproduction's gain or loss over the values reported by \citet{koreeda2021contractnli} (`K\&M'). O.SD and R.SD indicate whether the difference is less than or equal to the original standard deviations reported by the authors or the reproduction's standard deviations respectively.}
    \label{tab:reproduction-results}
\end{table*}
\begin{table*}[h]
\centering
{\small
\begin{tabularx}{0.8\textwidth}{l||ccc|c}
    \textbf{Model} & \textbf{Accuracy} & \textbf{F1(C)} & \textbf{F1(E)} & \textbf{Invalid Rate (\%)} \\
\hline
\hline
Majority (Ent.) & 0.463 & 0.000 & 0.590 & --- \\
\hline
Span NLI BERT &
    $\localbest{0.869\pm0.006}$ &
    $0.357\pm0.039$ &
    $\localbest{0.834\pm0.002}$ &
    ---\\
\hline
Gemma-4 E2B $^{\dag}$ &
    $0.688 \pm 0.003$ &
    $0.254 \pm 0.018$ &
    ${0.681 \pm 0.001}$ &
    $0.0 \pm 0.0$ \\
 Gemma-4 E4B $^{\dag}$&
    ${0.746 \pm 0.006}$ &
    $0.343 \pm 0.012$ &
    ${0.753 \pm 0.005}$ &
    $0.0 \pm 0.0$ \\
\rowcolor{SpringGreen}
 Gemma-4 26B $^{\ast}$&
    $0.812 \pm 0.002$ &
    $\localbest{0.443 \pm 0.030}$ &
    ${0.808 \pm 0.003}$ &
    $1.0 \pm 0.1$ \\
\hline
Ministral-3 3B $^{\dag}$ &
    ${0.575 \pm 0.003}$ &
    $0.110 \pm 0.019$ &
    $0.628 \pm 0.001$ &
    $0.0 \pm 0.0$ \\
Ministral-3 8B $^{\dag}$&
    $0.682 \pm 0.008$ &
    $0.325 \pm 0.045$ &
    $0.691 \pm 0.009$ &
    $0.0 \pm 0.0$ \\
 Ministral-3 14B $^{\dag}$&
    $0.720 \pm 0.003$ &
    $0.372 \pm 0.008$ &
    $0.728 \pm 0.002$ &
    $0.0 \pm 0.0$ \\
\hline
Qwen-3.5 4B $^{\dag}$&
    ${0.741 \pm 0.007}$ &
    $0.370 \pm 0.025$ &
    ${0.742 \pm 0.006}$ &
    $0.1 \pm 0.1$ \\
Qwen-3.5 9B $^{\dag}$&
    $0.759 \pm 0.004$ &
    $0.330 \pm 0.002$ &
    $0.749 \pm 0.002$ &
    $0.0 \pm 0.0$ \\
 Qwen-3.5 27B $^{\dag}$$^{\ast}$ &
    $0.793 \pm 0.005$ &
    $0.422 \pm 0.024$ &
    $0.783 \pm 0.006$ &
    $0.3 \pm 0.0$ \\
\hline
 Qwen-3.6 35B $^{\dag}$$^{\ast}$&
    ${0.760 \pm 0.006}$ &
    $0.330 \pm 0.005$ &
    $0.780 \pm 0.008$ &
    $0.6 \pm 0.2$ \\
  \end{tabularx}}\\
  {\small $^\ast$ Reasoning enabled \hspace{10mm}
  $^\dag$ Output structure applied}
  \caption{\small Results on three-way classification. Boldface indicates the best-performing model per metric, and the ``Invalid Rate (\%)'' column indicates the average percentage of answers outputted by the LLM from which we were unable to retrieve a coherent label. The green row highlights the best-performing LLM. }
  \label{tab:abridged-.3}
\end{table*}
\section{Results}

\subsection{Span NLI BERT reproduction}

The values we report in Table \ref{tab:reproduction-results} are very close to those reported by \citet{koreeda2021contractnli}. For all metrics except F1(E), our results are within one standard deviation from those reported by the authors. For F1(E), our reproduction scores 0.008 less than the values reported by Koreeda, a difference of about 1\%. As our reproduction has results very close to those reported by the authors on all metrics, we consider the method to have been successfully reproduced. 

\subsection{LLM evaluation}



\paragraph{ContractNLI} Our main results on three-way classification are in Table \ref{tab:abridged-.3}. 
We find that all models outperform the majority baseline. None of the models reach the performance of the fine-tuned Span NLI BERT on Accuracy and F1(E). However, several models match or surpass its performance on F1(C).
Of the tested LLMs, Gemma-4 26B performs best on all metrics, outperforming Span NLI BERT on F1(C) and outperforming all other models on other metrics.
Qwen-3.5 27B achieves a slightly worse, but still similar, performance to Gemma-4 26B. Qwen-3.6 35B performs noticeably worse than these two models on most metrics, but still outperforms several other models.
For medium-sized models (8B-14B parameters), Qwen-3.5 9B yields the best performance on overall accuracy and F1(E). On F1(C), it is outperformed by Ministral-3 14B. Ministral-3 8B overall has the lowest scores on metrics.
Among small models (2B-4B), Qwen-3.5 4B outperforms all other generative models on F1(C). Ministral-3 3B reports the lowest results out of all models, performing worse on all metrics compared to the similarly-sized Gemma-4 E2B and Qwen-3.5 4B. 
%

Invalid answer rates are very low across all models. All non-reasoning models have a 0.0\% invalid answer rate, with the exception of Qwen-3.5 4B, which has a 0.1\% invalid answer rate. Interestingly, the best performing model, Gemma-4 26B, has the highest invalid answer rate, with 1\% of its answers not containing a valid label.

\paragraph{Temperature analysis} The temperature analysis reveals that the tested models are largely robust against changes in temperature. Accuracy and F1(E) barely changes across all models, with F1(C) showing the most change. These changes can be attributed to the much smaller number of contradiction labels in the dataset. This imbalance would result in small changes in the number of correctly classified contradictions having a noticeable effect on F1(C). See Appendix \ref{app:temp-analysis} (Figure \ref{fig:temperature-analysis}) for detailed results.

\begin{table}[t]
\centering
\setlength{\tabcolsep}{4pt}
\newcommand{\pass}{\cellcolor{SpringGreen}$\checkmark$}
\newcommand{\fail}{$\times$}

\small
\begin{tabular}{l||cc|cc|cc
}
& \multicolumn{2}{c|}{\textbf{Entailment}} & \multicolumn{2}{c|}{\textbf{Contradiction}} & \multicolumn{2}{c}{\textbf{NotM.}} \\
\hline
\textbf{Model} & Avg & Max & Avg & Max & Avg & Max \\
\hline\hline
Gemma-4 E2B   & \pass & \pass & \pass & \pass & \pass & \pass \\
Gemma-4 E4B   & \pass & \pass & \pass & \pass & \pass & \pass \\
Gemma-4 26B   & \fail & \fail & \fail & \fail & \fail & \fail \\
\hline
Ministral-3 3B & \fail & \fail & \fail & \fail & \fail & \fail \\
Ministral-3 8B & \fail & \fail & \fail & \fail & \fail & \fail \\
Ministral-3 14B& \fail & \fail & \fail & \fail & \pass & \pass \\
\hline
Qwen-3.5 4B  & \pass & \pass & \fail & \fail & \pass & \pass \\
Qwen-3.5 9B  & \fail & \fail & \fail & \fail & \pass & \pass \\
Qwen-3.5 27B & \pass & \pass & \fail & \fail & \pass & \pass \\
\hline
Qwen-3.6 35B & \fail & \fail & \fail & \fail & \fail & \fail \\
\end{tabular}
\caption{\small Uncertainty analysis summary for the main results (Table \ref{tab:abridged-.3}) on the label-only scope. \textbf{{\checkmark}} indicates tests on all 3 seeds are significant ($\tilde{p}_i < 0.05$). $\times$ indicates otherwise.}
\label{tab:uq_summary}
\end{table}

\paragraph{Uncertainty analysis} 

The Mann-Whitney U Tests for the label-only analysis (Table \ref{tab:uq_summary}) indicate that, for most tested models, average negative log-probability and maximum negative log-probability do not reliably reflect answer correctness. The tests for Gemma-4 E2B and Gemma-4 E4B all have significant $\tilde{p}_i$-values for prediction token logits across all seeds and ground truth labels, suggesting that their output label logits are sufficiently aligned with predictive uncertainty. 

Some models have significant $\tilde{p}_i$-values on several tests, with Qwen-3.5 27B doing so on most label-only tests. However, it fails to achieve significant $\tilde{p}_i$-values across all seeds.

The results do not suggest a wider, generalizable trend between models of the same parameter counts or families.
Detailed results are available in Appendix \ref{app:logit-analysis-results}.

\paragraph{NLI4Wills} Our cross-domain evaluation on the NLI4Wills datasets shows that Qwen-3.6 35B is the best-performing LLM on these datasets. However, it falls short of the RoBERTa-based baseline results reported by \citet{kwak-et-al-nllp2023-error-analysis} on the Tennessee dataset. Detailed results are shown in Tables \ref{tab:nli4wills-idaho-results} and \ref{tab:nli4wills-tennessee-results} in Appendix~\ref{app:nli4wills-results}.

\paragraph{Ablation study} The best-performing settings per model are shown in Table \ref{tab:ablation-summary}. The ablations on Gemma-4 models do not have a large effect on the accuracy, but F1(C) scores on the smaller models tend to be higher for the reasoning-only settings. This difference is less noticeable for Gemma-4 26B, where scores between reasoning-only and structure-only are mostly the same.

Qwen-3.5 models tend to have a very high invalid output rate ($12-69\%$) when for the reasoning-only ablation. This results in a much lower model accuracy compared to other settings. The invalid output rate decreases greatly when both structures and reasoning are used, and becomes negligible ($\leq0.1\%$) for the structure-only ablation.
Qwen-3.6 35B performs best across all metrics in the reasoning-only ablation while maintaining a very low invalid output rate.

\begin{table}[h]
\centering
    \resizebox{.5\textwidth}{!}
{
\begin{tabular}{@{}l||ccc|c@{}}
    \textbf{Model} & \textbf{Accuracy} & \textbf{F1(C)} & \textbf{F1(E)} & \textbf{IR (\%)} \\
\hline
\hline
S-NLI-BERT &
    \textbf{0.869$_{0.006}$} &
    0.357$_{0.039}$ &
    \textbf{0.834$_{0.002}$} &
    ---\\
\hline
Gemma-4 E2B $^{\ast}$ &
         0.680$_{0.013}$ & {0.268$_{0.026}$} & {0.692$_{0.008}$}
         & 0.0$_{0.0}$\\
 Gemma-4 E4B $^{\ast}$ &
         0.744$_{0.004}$ & {0.386$_{0.013}$} & 0.747$_{0.004}$
         & 0.0$_{0.0}$ \\
 \rowcolor{SpringGreen}  Gemma-4 26B$^{\dag}$&
    {0.827$_{0.003}$} & \textbf{0.456$_{0.016}$} & 0.803$_{0.001}$
         & 0.0$_{0.0}$  \\
\hline
Qwen-3.5 4B $^{\dag}$ &
    0.741$_{0.007}$ & {0.370$_{0.025}$} & {0.742$_{0.006}$}
    & 0.1$_{0.1}$ \\
Qwen-3.5 9B $^{\dag}$ &
    {0.759$_{0.004}$} & 0.330$_{0.002}$ & {0.749$_{0.002}$} & 0.0$_{0.0}$    \\
 Qwen-3.5 27B$^{\dag\ast}$ &
    0.793$_{0.005}$ & {0.422$_{0.024}$} & 0.783$_{0.006}$ & 0.3$_{0.0}$    \\
\hline
 Qwen-3.6 35B$^{\ast}$&
    {0.791$_{0.005}$} & {0.387$_{0.021}$} & {0.793$_{0.003}$}
         & 0.1$_{0.1}$  \\
  \end{tabular}}\\
  {\small $^\ast$ Reasoning enabled \hfill
  $^\dag$ Output structure applied}
  \caption{\small Ablation study -- best setting per model. Boldface indicates the best-performing model per metric. The IR (\%) column corresponds to the invalid answer rate. Subscripts correspond to standard deviations. The green row highlights the best-performing LLM.}
  \label{tab:ablation-summary}
\end{table}

Across all models, using structured outputs and reasoning simultaneously rarely results in higher scores compared to using only one of the two, nor does it result in lower invalid output rates. Detailed results are available in Appendix \ref{app:class-specific-metrics}.



\section{Conclusion}


        In this paper, we successfully reproduced Span NLI BERT, and investigated the performance of ten zero-shot LLMs in three model families on the NLI task of the ContractNLI dataset.
We found that the best performing LLM is Gemma-4 26B with an accuracy of 81.2\%, scoring 5.7\% below the supervised model. Qwen-3.6 35B shows promising results on both the ContractNLI and NLI4Wills datasets. 
This indicates that although zero-shot open-source generative quantized LLMs underperform fine-tuned models on ContractNLI, they are a viable alternative in settings without any labelled data.
As next steps, we plan to bring our models to the inspectorates' work practice to evaluate them for other languages and subdomains. 
Additionally, we will experiment with UQ methods, including the use of reasoning capabilities to provide explanations for the generated label.

\section*{Acknowledgments}

This research is funded by the ICAI lab AI4Oversight.

We thank SURF (\url{www.surf.nl}) for the support in using the National Supercomputer Snellius.

This work was partially performed using the compute resources from the Academic Leiden Interdisciplinary Cluster Environment (ALICE) provided by Leiden University.

\section*{Limitations}

We test a good number of LLMs, all quantized via Ollama, but adding more models is always possible. Therefore, the results we present may not be representative of \textit{all} quantized models' performance on this task.


Our study mostly uses a fixed configuration for variables such as reasoning modes, structures, and prompts. Therefore, our results are non-exhaustive, and we do not make claims on the performance of our setup compared to alternate configurations for the same models.

While our experimental procedures and temperature settings prevent label predictions generated by the LLM from being misaligned from the generated explanations, the stochastic nature of such models means that this is still a small possibility. As a result, different seeds and temperatures may result in small variations in classification even if the input and explanation remain the same. We expect that this does not have a meaningful effect on results.

We have not been able to find any legal NLI datasets with documents longer than ContractNLI, which has a maximum document length of about 12,000 tokens. All datasets used contain inputs that fit comfortably into our LLMs' context windows, and using methods for handling very lengthy documents was not necessary for our experiments. Therefore, our results are not representative of these models' performance on very lengthy documents that exceed their context windows, or their performance when combined with methods explicitly designed for handling very long documents.
\section*{Ethical considerations}
Our work is intended to assist experts in their decision-making processes by providing them with helpful tools. We do not seek to design systems that replace experts, especially given the risky nature of the regulatory domain. We maintain regular communication with several inspectorates, and ensure that the experts and the inspectorates as a whole are aware of this.

\bibliography{custom}

\newpage

\appendix

\section{Prompts}\label{app:prompts}

\subsection{ContractNLI Prompt}
\begin{lstlisting}
Document: {context_text}

Hypothesis: {hypothesis_text}

Given a document and a hypothesis, determine whether the document entails, contradicts, or is unrelated to the hypothesis. Be concise, and limit your answer to no more than 200 words. Answer strictly as "Entailment", "Contradiction", or "NotMentioned".
\end{lstlisting}

\subsection{NLI4Wills Prompt}
\begin{lstlisting}
Law: {law_text}

Conditions: {conditions_text}

Statement: {statement_text}


Given a law, conditions, and statement, determine whether the law and condition refute, support, or are unrelated to the statement. Be concise, and limit your answer to no more than 200 words. Answer strictly as "Refute", "Support", or "Unrelated".
\end{lstlisting}








\section{Structures}\label{app:structures}
Structures are implemented in Python using the \texttt{Pydantic} package \texttt{BaseModel} class.
\subsection{Main Structure}
\begin{lstlisting}
{
    explanation: str
    classification: [
        "Contradiction", 
        "Entailment", 
        "NotMentioned"
    ]
}
\end{lstlisting}



\subsection{NLI4Wills Structure}\label{sec:nli4wills-struct}
\begin{lstlisting}
{
    explanation: str
    classification: [
        "support", 
        "refute",
        "unrelated"
    ]
}
\end{lstlisting}

\section{Uncertainty Quantification Metrics}\label{app:uq-equations}
All equations are taken from the survey work by \citet{shorinwa2025survey}. See Footnote \ref{foot:shorinwa} for details.
\paragraph{Maximum Negative Token Log-Probability}

\begin{equation}
\text{Maximum}(p) = \max_j(-\log(p_{ij}))
\end{equation}

\paragraph{Average Negative Token Log-Probability}

\begin{equation}
\text{Average}(p) = -\frac{1}{L_i} \sum_j \log(p_{ij})
\end{equation}



\section{Hardware}

\paragraph{Span NLI Reproduction} In order to closely replicate the author's hardware, we train the model on Nvidia A100 GPUs. We use four GPUs instead of eight. 

\paragraph{Other experiments} We ran our experiments on a variety of hardware, with availability being the main factor in hardware selection. The following GPU models were used:
\begin{itemize}
    \item Nvidia RTX 5070 Ti
    \item Nvidia RTX 3090
    \item Nvidia L4
    \item Nvidia A100
    \item Nvidia A100 MIG (40GB instances)
\end{itemize}

\clearpage


\begin{figure*}[h]
    \section{Temperature analysis} \label{app:temp-analysis}
    \centering

    \begin{subfigure}[t]{0.25\textwidth}
    \begin{tikzpicture}
    \begin{axis}[
        title        = {Gemma-4},
        xlabel       = {Temperature},
        ylabel       = {Accuracy},
        legend pos   = south west,
        legend style = {font=\small},
        ymin = 0, ymax = 1.0,
        grid=major, grid style={dashed,gray!40},
        width=5.2cm, height=5cm,
        xtick={0.1, 0.2, 0.3, 0.4, 0.5},
        thick,
    ]
        \addplot table [x index=0, y index=1] {inserts/temp-graphs/gemma4.txt};
        \addlegendentry{E2B}
        \addplot table [x index=0, y index=3] {inserts/temp-graphs/gemma4.txt};
        \addlegendentry{E4B}
        \addplot table [x index=0, y index=2] {inserts/temp-graphs/gemma4.txt};
        \addlegendentry{26B}
    \end{axis}
    \end{tikzpicture}
    \end{subfigure}%
    \hfill%
    \begin{subfigure}[t]{0.25\textwidth}
    \begin{tikzpicture}
    \begin{axis}[
        title        = {Ministral-3},
        xlabel       = {Temperature},
        legend pos   = south west,
        legend style = {font=\small},
        ymin = 0, ymax = 1.0,
        grid=major, grid style={dashed,gray!40},
        width=5.2cm, height=5cm,
        xtick={0.1, 0.2, 0.3, 0.4, 0.5},
        thick,
    ]
        \addplot table [x index=0, y index=3] {inserts/temp-graphs/ministral-3.txt};
        \addlegendentry{3B}
        \addplot table [x index=0, y index=2] {inserts/temp-graphs/ministral-3.txt};
        \addlegendentry{8B}
        \addplot table [x index=0, y index=1] {inserts/temp-graphs/ministral-3.txt};
        \addlegendentry{14B}
    \end{axis}
    \end{tikzpicture}
    \end{subfigure}%
    \hfill%
    \begin{subfigure}[t]{0.25\textwidth}
    \begin{tikzpicture}
    \begin{axis}[
        title        = {Qwen-3.5/3.6},
        xlabel       = {Temperature},
        legend pos   = south west,
        legend style = {font=\small},
        ymin = 0, ymax = 1.0,
        grid=major, grid style={dashed,gray!40},
        width=5.2cm, height=5cm,
        xtick={0.1, 0.2, 0.3, 0.4, 0.5},
        thick,
    ]
        \addplot table [x index=0, y index=1] {inserts/temp-graphs/qwen35.txt};
        \addlegendentry{3.5/4B}
        \addplot table [x index=0, y index=4] {inserts/temp-graphs/qwen35.txt};
        \addlegendentry{3.5/9B}
        \addplot table [x index=0, y index=3] {inserts/temp-graphs/qwen35.txt};
        \addlegendentry{3.5/27B}
        \addplot table [x index=0, y index=1] {inserts/temp-graphs/qwen36.txt};
        \addlegendentry{3.6/35B}
    \end{axis}
    \end{tikzpicture}
    \end{subfigure}

    \begin{subfigure}[t]{0.25\textwidth}
    \begin{tikzpicture}
    \begin{axis}[
        xlabel       = {Temperature},
        ylabel       = {F1(C)},
        legend pos   = north west,
        legend style = {font=\small},
        ymin = 0, ymax = 1.0,
        grid=major, grid style={dashed,gray!40},
        width=5.2cm, height=5cm,
        xtick={0.1, 0.2, 0.3, 0.4, 0.5},
        thick,
    ]
        \addplot table [x index=0, y index=1] {inserts/temp-graphs/f1c/gemma4.txt};
        \addlegendentry{E2B}
        \addplot table [x index=0, y index=3] {inserts/temp-graphs/f1c/gemma4.txt};
        \addlegendentry{E4B}
        \addplot table [x index=0, y index=2] {inserts/temp-graphs/f1c/gemma4.txt};
        \addlegendentry{26B}
    \end{axis}
    \end{tikzpicture}
    \end{subfigure}%
    \hfill%
    \begin{subfigure}[t]{0.25\textwidth}
    \begin{tikzpicture}
    \begin{axis}[
        xlabel       = {Temperature},
        legend pos   = north west,
        legend style = {font=\small},
        ymin = 0, ymax = 1.0,
        grid=major, grid style={dashed,gray!40},
        width=5.2cm, height=5cm,
        xtick={0.1, 0.2, 0.3, 0.4, 0.5},
        thick,
    ]
        \addplot table [x index=0, y index=3] {inserts/temp-graphs/f1c/ministral-3.txt};
        \addlegendentry{3B}
        \addplot table [x index=0, y index=2] {inserts/temp-graphs/f1c/ministral-3.txt};
        \addlegendentry{8B}
        \addplot table [x index=0, y index=1] {inserts/temp-graphs/f1c/ministral-3.txt};
        \addlegendentry{14B}
    \end{axis}
    \end{tikzpicture}
    \end{subfigure}%
    \hfill%
    \begin{subfigure}[t]{0.25\textwidth}
    \begin{tikzpicture}
    \begin{axis}[
        xlabel       = {Temperature},
        legend pos   = north west,
        legend style = {font=\small, nodes={scale=0.7, transform shape}},
        ymin = 0, ymax = 1.0,
        grid=major, grid style={dashed,gray!40},
        width=5.2cm, height=5cm,
        xtick={0.1, 0.2, 0.3, 0.4, 0.5},
        thick,
    ]
        \addplot table [x index=0, y index=1] {inserts/temp-graphs/f1c/qwen35.txt};
        \addlegendentry{3.5/4B}
        \addplot table [x index=0, y index=4] {inserts/temp-graphs/f1c/qwen35.txt};
        \addlegendentry{3.5/9B}
        \addplot table [x index=0, y index=3] {inserts/temp-graphs/f1c/qwen35.txt};
        \addlegendentry{3.5/27B}
        \addplot table [x index=0, y index=1] {inserts/temp-graphs/f1c/qwen36.txt};
        \addlegendentry{3.6/35B}
    \end{axis}
    \end{tikzpicture}
    \end{subfigure}

    \begin{subfigure}[t]{0.25\textwidth}
    \begin{tikzpicture}
    \begin{axis}[
        xlabel       = {Temperature},
        ylabel       = {F1(E)},
        legend pos   = south west,
        legend style = {font=\small},
        ymin = 0, ymax = 1.0,
        grid=major, grid style={dashed,gray!40},
        width=5.2cm, height=5cm,
        xtick={0.1, 0.2, 0.3, 0.4, 0.5},
        thick,
    ]
        \addplot table [x index=0, y index=1] {inserts/temp-graphs/f1e/gemma4.txt};
        \addlegendentry{E2B}
        \addplot table [x index=0, y index=3] {inserts/temp-graphs/f1e/gemma4.txt};
        \addlegendentry{E4B}
        \addplot table [x index=0, y index=2] {inserts/temp-graphs/f1e/gemma4.txt};
        \addlegendentry{26B}
    \end{axis}
    \end{tikzpicture}
    \end{subfigure}%
    \hfill%
    \begin{subfigure}[t]{0.25\textwidth}
    \begin{tikzpicture}
    \begin{axis}[
        xlabel       = {Temperature},
        legend pos   = south west,
        legend style = {font=\small},
        ymin = 0, ymax = 1.0,
        grid=major, grid style={dashed,gray!40},
        width=5.2cm, height=5cm,
        xtick={0.1, 0.2, 0.3, 0.4, 0.5},
        thick,
    ]
        \addplot table [x index=0, y index=3] {inserts/temp-graphs/f1e/ministral-3.txt};
        \addlegendentry{3B}
        \addplot table [x index=0, y index=2] {inserts/temp-graphs/f1e/ministral-3.txt};
        \addlegendentry{8B}
        \addplot table [x index=0, y index=1] {inserts/temp-graphs/f1e/ministral-3.txt};
        \addlegendentry{14B}
    \end{axis}
    \end{tikzpicture}
    \end{subfigure}%
    \hfill%
    \begin{subfigure}[t]{0.25\textwidth}
    \begin{tikzpicture}
    \begin{axis}[
        xlabel       = {Temperature},
        legend pos   = south west,
        legend style = {font=\small},
        ymin = 0, ymax = 1.0,
        grid=major, grid style={dashed,gray!40},
        width=5.2cm, height=5cm,
        xtick={0.1, 0.2, 0.3, 0.4, 0.5},
        thick,
    ]
        \addplot table [x index=0, y index=1] {inserts/temp-graphs/f1e/qwen35.txt};
        \addlegendentry{3.5/4B}
        \addplot table [x index=0, y index=4] {inserts/temp-graphs/f1e/qwen35.txt};
        \addlegendentry{3.5/9B}
        \addplot table [x index=0, y index=3] {inserts/temp-graphs/f1e/qwen35.txt};
        \addlegendentry{3.5/27B}
        \addplot table [x index=0, y index=1] {inserts/temp-graphs/f1e/qwen36.txt};
        \addlegendentry{3.6/35B}
    \end{axis}
    \end{tikzpicture}
    \end{subfigure}

    \caption{Effect of temperature on ContractNLI results grouped by model family.}

    \label{fig:temperature-analysis}

\end{figure*}
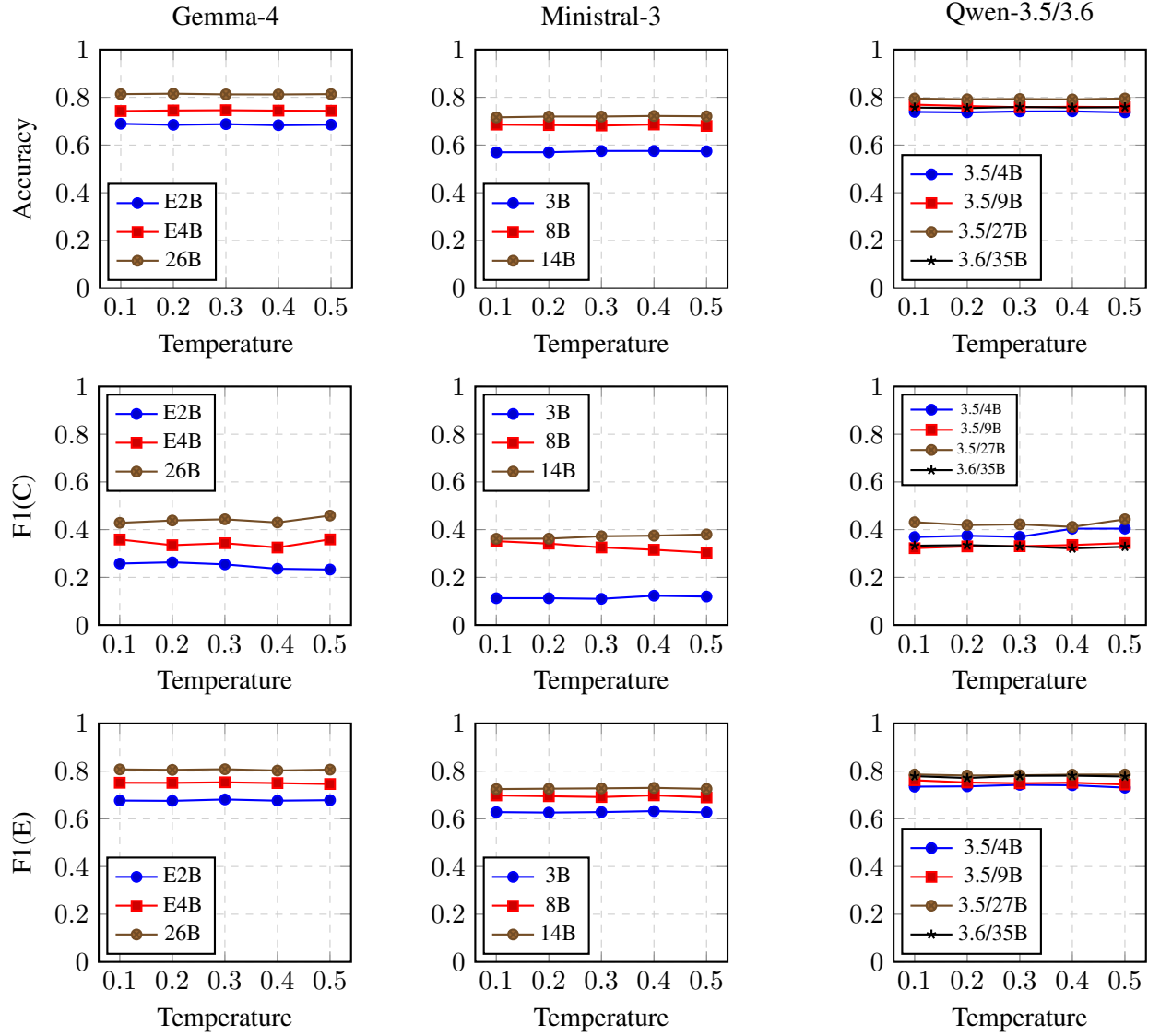

\begin{table*}[h]
  \centering
  \small
  \begin{tabularx}{0.865\textwidth}{l|l|lll|l}
    \textbf{Model} & \textbf{Temp} & \textbf{Accuracy} & \textbf{F1(C)} & \textbf{F1(E)} & \textbf{Invalid Rate (\%)} \\
    \hline
    \hline
Gemma-4 E2B $^{\dag}$& 0.1 & 
    $\mathbf{0.689 \pm 0.002}$ &
    $0.258 \pm 0.018$ &
    $0.676 \pm 0.004$ &
    $0.000 \pm 0.000$ \\
\rowcolor{SpringGreen}
 & 0.2 &
    $0.685 \pm 0.001$ &
    $\mathbf{0.263 \pm 0.006}$ &
    $0.675 \pm 0.002$ &
    $0.000 \pm 0.000$ \\
 & 0.3 &
    $0.688 \pm 0.003$ &
    $0.254 \pm 0.018$ &
    $\mathbf{0.681 \pm 0.001}$ &
    $0.016 \pm 0.023$ \\
 & 0.4 &
    $0.683 \pm 0.008$ &
    $0.236 \pm 0.011$ &
    $0.676 \pm 0.006$ &
    $0.016 \pm 0.023$ \\
 & 0.5 &
    $0.685 \pm 0.007$ &
    $0.233 \pm 0.003$ &
    $0.678 \pm 0.006$ &
    $0.000 \pm 0.000$ \\
\hline
Gemma-4 E4B $^{\dag}$ & 0.1 &
    $0.743 \pm 0.004$ &
    $\mathbf{0.359 \pm 0.040}$ &
    $0.751 \pm 0.002$ &
    $0.000 \pm 0.000$ \\
 & 0.2 &
    $0.745 \pm 0.005$ &
    $0.335 \pm 0.007$ &
    $0.751 \pm 0.006$ &
    $0.000 \pm 0.000$ \\
 & 0.3 &
    $\mathbf{0.746 \pm 0.006}$ &
    $0.343 \pm 0.012$ &
    $\mathbf{0.753 \pm 0.005}$ &
    $0.000 \pm 0.000$ \\
 & 0.4 &
    $0.744 \pm 0.003$ &
    $0.325 \pm 0.003$ &
    $0.750 \pm 0.002$ &
    $0.000 \pm 0.000$ \\
\rowcolor{SpringGreen}
 & 0.5 &
    $0.744 \pm 0.005$ &
    $\mathbf{0.359 \pm 0.018}$ &
    $0.746 \pm 0.005$ &
    $0.000 \pm 0.000$ \\
\hline
Gemma-4 26B $^{\ast }$ & 0.1 &
    $0.813 \pm 0.002$ &
    $0.428 \pm 0.022$ &
    $0.807 \pm 0.002$ &
    $1.004 \pm 0.078$ \\
 & 0.2 &
    $\mathbf{0.815 \pm 0.002}$ &
    $0.438 \pm 0.027$ &
    $0.805 \pm 0.002$ &
    $0.861 \pm 0.170$ \\
 & 0.3 &
    $0.812 \pm 0.002$ &
    $0.443 \pm 0.030$ &
    $\mathbf{0.808 \pm 0.003}$ &
    $0.972 \pm 0.148$ \\
 & 0.4 &
    $0.812 \pm 0.001$ &
    $0.430 \pm 0.018$ &
    $0.802 \pm 0.001$ &
    $0.733 \pm 0.180$ \\
\rowcolor{SpringGreen}
 & 0.5 &
    $0.813 \pm 0.002$ &
    $\mathbf{0.459 \pm 0.022}$ &
    $0.806 \pm 0.001$ &
    $0.956 \pm 0.217$ \\
\hline
\hline
Ministral-3 3B $^{\dag}$ & 0.1 &
    $0.570 \pm 0.002$ &
    $0.113 \pm 0.011$ &
    $0.628 \pm 0.004$ &
    $0.000 \pm 0.000$ \\
 & 0.2 &
    $0.570 \pm 0.005$ &
    $0.113 \pm 0.011$ &
    $0.626 \pm 0.001$ &
    $0.000 \pm 0.000$ \\
 & 0.3 &
    $\mathbf{0.575 \pm 0.003}$ &
    $0.110 \pm 0.019$ &
    $0.628 \pm 0.001$ &
    $0.000 \pm 0.000$ \\
\rowcolor{SpringGreen}
 & 0.4 &
    $\mathbf{0.575 \pm 0.005}$ &
    $\mathbf{0.123 \pm 0.014}$ &
    $\mathbf{0.632 \pm 0.003}$ &
    $0.000 \pm 0.000$ \\
 & 0.5 &
    $0.574 \pm 0.006$ &
    $0.120 \pm 0.029$ &
    $0.627 \pm 0.004$ &
    $0.000 \pm 0.000$ \\
\hline
\rowcolor{SpringGreen}
Ministral-3 8B $^{\dag}$ & 0.1 &
    $\mathbf{0.686 \pm 0.003}$ &
    $\mathbf{0.352 \pm 0.016}$ &
    $\mathbf{0.698 \pm 0.007}$ &
    $0.000 \pm 0.000$ \\
 & 0.2 &
    $0.684 \pm 0.000$ &
    $0.341 \pm 0.059$ &
    $0.695 \pm 0.009$ &
    $0.000 \pm 0.000$ \\
 & 0.3 &
    $0.682 \pm 0.008$ &
    $0.325 \pm 0.045$ &
    $0.691 \pm 0.009$ &
    $0.000 \pm 0.000$ \\
 & 0.4 &
    $\mathbf{0.686 \pm 0.003}$ &
    $0.316 \pm 0.026$ &
    $\mathbf{0.698 \pm 0.008}$ &
    $0.000 \pm 0.000$ \\
 & 0.5 &
    $0.680 \pm 0.001$ &
    $0.303 \pm 0.018$ &
    $0.690 \pm 0.002$ &
    $0.000 \pm 0.000$ \\
\hline
Ministral-3 14B $^{\dag}$ & 0.1 &
    $0.716 \pm 0.005$ &
    $0.362 \pm 0.019$ &
    $0.725 \pm 0.009$ &
    $0.000 \pm 0.000$ \\
 & 0.2 &
    $0.720 \pm 0.003$ &
    $0.362 \pm 0.028$ &
    $0.726 \pm 0.004$ &
    $0.000 \pm 0.000$ \\
 & 0.3 &
    $0.720 \pm 0.003$ &
    $0.372 \pm 0.008$ &
    $0.728 \pm 0.002$ &
    $0.000 \pm 0.000$ \\
\rowcolor{SpringGreen}
 & 0.4 &
    $\mathbf{0.722 \pm 0.006}$ &
    $0.375 \pm 0.010$ &
    $\mathbf{0.730 \pm 0.004}$ &
    $0.000 \pm 0.000$ \\
 & 0.5 &
    $0.721 \pm 0.003$ &
    $\mathbf{0.380 \pm 0.019}$ &
    $0.726 \pm 0.003$ &
    $0.000 \pm 0.000$ \\
\hline
\hline
Qwen-3.5 4B $^{\dag}$ & 0.1 &
    $0.739 \pm 0.006$ &
    $0.369 \pm 0.011$ &
    $0.735 \pm 0.007$ &
    $0.080 \pm 0.081$ \\
 & 0.2 &
    $0.737 \pm 0.004$ &
    $0.374 \pm 0.022$ &
    $0.736 \pm 0.002$ &
    $0.159 \pm 0.081$ \\
 & 0.3 &
    $\mathbf{0.741 \pm 0.007}$ &
    $0.370 \pm 0.025$ &
    $\mathbf{0.742 \pm 0.006}$ &
    $0.112 \pm 0.060$ \\
\rowcolor{SpringGreen}
 & 0.4 &
    $\mathbf{0.741 \pm 0.006}$ &
    $\mathbf{0.404 \pm 0.005}$ &
    $0.741 \pm 0.007$ &
    $0.112 \pm 0.060$ \\
 & 0.5 &
    $0.737 \pm 0.007$ &
    $\mathbf{0.404 \pm 0.026}$ &
    $0.731 \pm 0.011$ &
    $0.112 \pm 0.081$ \\
\hline
\rowcolor{SpringGreen}
Qwen-3.5 9B $^{\dag}$ & 0.1 &
    $\mathbf{0.769 \pm 0.004}$ &
    $0.322 \pm 0.014$ &
    $\mathbf{0.762 \pm 0.004}$ &
    $0.016 \pm 0.023$ \\
 & 0.2 &
    $0.763 \pm 0.002$ &
    $0.330 \pm 0.011$ &
    $0.752 \pm 0.002$ &
    $0.032 \pm 0.023$ \\
 & 0.3 &
    $0.759 \pm 0.004$ &
    $0.330 \pm 0.002$ &
    $0.749 \pm 0.002$ &
    $0.016 \pm 0.023$ \\
 & 0.4 &
    $0.760 \pm 0.010$ &
    $0.335 \pm 0.051$ &
    $0.751 \pm 0.008$ &
    $0.016 \pm 0.023$ \\
 & 0.5 &
    $0.759 \pm 0.006$ &
    $\mathbf{0.344 \pm 0.038}$ &
    $0.744 \pm 0.008$ &
    $0.016 \pm 0.023$ \\
\hline
Qwen-3.5 27B $^{\dag\ast}$ & 0.1 &
    $\mathbf{0.795 \pm 0.005}$ &
    $0.431 \pm 0.018$ &
    $\mathbf{0.786 \pm 0.007}$ &
    $0.351 \pm 0.137$ \\
 & 0.2 &
    $0.792 \pm 0.002$ &
    $0.419 \pm 0.024$ &
    $0.782 \pm 0.006$ &
    $0.351 \pm 0.081$ \\
 & 0.3 &
    $0.793 \pm 0.005$ &
    $0.422 \pm 0.024$ &
    $0.783 \pm 0.006$ &
    $0.319 \pm 0.060$ \\
 & 0.4 &
    $0.791 \pm 0.005$ &
    $0.412 \pm 0.030$ &
    $0.785 \pm 0.006$ &
    $0.239 \pm 0.068$ \\
\rowcolor{SpringGreen}
 & 0.5 &
    $\mathbf{0.795 \pm 0.003}$ &
    $\mathbf{0.443 \pm 0.011}$ &
    $\mathbf{0.786 \pm 0.002}$ &
    $0.223 \pm 0.090$ \\
\hline
\hline
Qwen-3.6 35B $^{\dag\ast}$ & 0.1 &
    $0.757 \pm 0.001$ &
    $0.333 \pm 0.004$ &
    $0.779 \pm 0.002$ &
    $0.670 \pm 0.141$ \\
 & 0.2 &
    $0.755 \pm 0.003$ &
    $\mathbf{0.335 \pm 0.002}$ &
    $0.771 \pm 0.003$ &
    $0.701 \pm 0.060$ \\
\rowcolor{SpringGreen}
 & 0.3 &
    $\mathbf{0.760 \pm 0.006}$ &
    $0.330 \pm 0.005$ &
    $0.780 \pm 0.008$ &
    $0.622 \pm 0.207$ \\
 & 0.4 &
    $0.757 \pm 0.004$ &
    $0.321 \pm 0.014$ &
    $\mathbf{0.781 \pm 0.007}$ &
    $0.574 \pm 0.103$ \\
 & 0.5 &
    $0.759 \pm 0.002$ &
    $0.328 \pm 0.010$ &
    $0.778 \pm 0.006$ &
    $0.542 \pm 0.126$ \\
  \end{tabularx}\\
  {\small $^\ast$ Reasoning enabled \hspace{20mm}
  $^\dag$ Output structure applied}
  \caption{Temperature analysis results. Each metric reports the average and standard deviation across three seeds. Boldface indicates the best-performing model per metric, and the ``Invalid Rate (\%)'' column indicates the average percentage of answers outputted by the LLM from which we were unable to retrieve a coherent label. The green row highlights the best-performing temperature setting per LLM.}
  \label{tab:tempsweep}

\end{table*}

\clearpage
\begin{table*}[htbp]
\section{Logit analysis results}\label{app:logit-analysis-results}
\subsection{Gemma-4 E2B}
\centering\begin{subtable}{0.48\textwidth}
\begin{tabular}{l|c||c|c}
\textbf{Label} & \textbf{Seed} & $\tilde{p}_i$ & \textbf{Effect size} \\
\hline
\hline

\multirow{3}{*}{Ent}
	 & 0 &  1.0 & -0.207 \\
	 & 1 &  1.0 & -0.19 \\
	 & 42 &  1.0 & -0.154 \\
\hline 
 \multirow{3}{*}{Cont}
	 & 0 &  0.478 & 0.083 \\
	 & 1 &  0.435 & 0.093 \\
	 & 42 &  0.684 & 0.056 \\
\hline 
 \multirow{3}{*}{NotM}
	 & 0 &  1.0 & -0.035 \\
	 & 1 &  1.0 & -0.044 \\
	 & 42 &  1.0 & -0.116 \\
\hline 
 \end{tabular} \caption{Average negative log-probability}\end{subtable}\begin{subtable}{0.48\textwidth}
\begin{tabular}{l|c||c|c}
\textbf{Label} & \textbf{Seed} & $\tilde{p}_i$ & \textbf{Effect size} \\
\hline
\hline

\multirow{3}{*}{Ent}
	 & 0 &  1.0 & -0.406 \\
	 & 1 &  1.0 & -0.4 \\
	 & 42 &  1.0 & -0.396 \\
\hline 
 \multirow{3}{*}{Cont}
	 & 0 &  1.0 & -0.318 \\
	 & 1 &  1.0 & -0.235 \\
	 & 42 &  1.0 & -0.208 \\
\hline 
 \multirow{3}{*}{NotM}
	 & 0 & \cellcolor{SpringGreen}  $\tilde{p}_i < 0.01$ & 0.379 \\
	 & 1 & \cellcolor{SpringGreen}  $\tilde{p}_i < 0.01$ & 0.413 \\
	 & 42 & \cellcolor{SpringGreen}  $\tilde{p}_i < 0.01$ & 0.445 \\
\hline 
 \end{tabular} \caption{Maximum negative token log-probability}\end{subtable}\caption{Gemma-4 E2B Mann-Whitney U Test results and effect sizes for uncertainty metrics calculated with \textbf{logits of the whole response}.}\end{table*}

 \begin{table*}[htbp] \centering\begin{subtable}{0.48\textwidth}
\begin{tabular}{l|c||c|c}
Label & Seed & $\tilde{p}_i$ & Effect size \\
\hline
\hline

\multirow{3}{*}{Ent}
	 & 0 & \cellcolor{SpringGreen}  $\tilde{p}_i < 0.01$ & 0.493 \\
	 & 1 & \cellcolor{SpringGreen}  $\tilde{p}_i < 0.01$ & 0.501 \\
	 & 42 & \cellcolor{SpringGreen}  $\tilde{p}_i < 0.01$ & 0.539 \\
\hline 
 \multirow{3}{*}{Cont}
	 & 0 & \cellcolor{SpringGreen}  $\tilde{p}_i < 0.01$ & 0.375 \\
	 & 1 & \cellcolor{SpringGreen}  $\tilde{p}_i < 0.01$ & 0.451 \\
	 & 42 & \cellcolor{SpringGreen}  $\tilde{p}_i < 0.01$ & 0.423 \\
\hline 
 \multirow{3}{*}{NotM}
	 & 0 & \cellcolor{SpringGreen}  $\tilde{p}_i < 0.01$ & 0.505 \\
	 & 1 & \cellcolor{SpringGreen}  $\tilde{p}_i < 0.01$ & 0.524 \\
	 & 42 & \cellcolor{SpringGreen}  $\tilde{p}_i < 0.01$ & 0.583 \\
\hline 
 \end{tabular} \caption{Average negative log-probability}\end{subtable}\begin{subtable}{0.48\textwidth}
\begin{tabular}{l|c||c|c}
Label & Seed & $\tilde{p}_i$ & Effect size \\
\hline
\hline

\multirow{3}{*}{Ent}
	 & 0 & \cellcolor{SpringGreen}  $\tilde{p}_i < 0.01$ & 0.493 \\
	 & 1 & \cellcolor{SpringGreen}  $\tilde{p}_i < 0.01$ & 0.501 \\
	 & 42 & \cellcolor{SpringGreen}  $\tilde{p}_i < 0.01$ & 0.539 \\
\hline 
 \multirow{3}{*}{Cont}
	 & 0 & \cellcolor{SpringGreen}  $\tilde{p}_i < 0.01$ & 0.373 \\
	 & 1 & \cellcolor{SpringGreen}  $\tilde{p}_i < 0.01$ & 0.448 \\
	 & 42 & \cellcolor{SpringGreen}  $\tilde{p}_i < 0.01$ & 0.42 \\
\hline 
 \multirow{3}{*}{NotM}
	 & 0 & \cellcolor{SpringGreen}  $\tilde{p}_i < 0.01$ & 0.505 \\
	 & 1 & \cellcolor{SpringGreen}  $\tilde{p}_i < 0.01$ & 0.524 \\
	 & 42 & \cellcolor{SpringGreen}  $\tilde{p}_i < 0.01$ & 0.583 \\
\hline 
 \end{tabular} \caption{Maximum negative token log-probability}\end{subtable}\caption{Gemma-4 E2B  Mann-Whitney U Test results and effect sizes for uncertainty metrics calculated with \textbf{only the label logits}.}\end{table*}

 \begin{table*}[htbp] \centering\begin{subtable}{0.48\textwidth}
\begin{tabular}{l|c||c|c}
Label & Seed & $\tilde{p}_i$ & Effect size \\
\hline
\hline

\multirow{3}{*}{Ent}
	 & 0 & \cellcolor{SpringGreen}  $\tilde{p}_i < 0.01$ & 0.493 \\
	 & 1 & \cellcolor{SpringGreen}  $\tilde{p}_i < 0.01$ & 0.501 \\
	 & 42 & \cellcolor{SpringGreen}  $\tilde{p}_i < 0.01$ & 0.539 \\
\hline 
 \multirow{3}{*}{Cont}
	 & 0 & \cellcolor{SpringGreen}  $\tilde{p}_i < 0.01$ & 0.365 \\
	 & 1 & \cellcolor{SpringGreen}  $\tilde{p}_i < 0.01$ & 0.44 \\
	 & 42 & \cellcolor{SpringGreen}  $\tilde{p}_i < 0.01$ & 0.416 \\
\hline 
 \multirow{3}{*}{NotM}
	 & 0 & \cellcolor{SpringGreen}  $\tilde{p}_i < 0.01$ & 0.505 \\
	 & 1 & \cellcolor{SpringGreen}  $\tilde{p}_i < 0.01$ & 0.524 \\
	 & 42 & \cellcolor{SpringGreen}  $\tilde{p}_i < 0.01$ & 0.583 \\
\hline 
 \end{tabular} \caption{Average negative log-probability}\end{subtable}\begin{subtable}{0.48\textwidth}
\begin{tabular}{l|c||c|c}
Label & Seed & $\tilde{p}_i$ & Effect size \\
\hline
\hline

\multirow{3}{*}{Ent}
	 & 0 & \cellcolor{SpringGreen}  $\tilde{p}_i < 0.01$ & 0.493 \\
	 & 1 & \cellcolor{SpringGreen}  $\tilde{p}_i < 0.01$ & 0.501 \\
	 & 42 & \cellcolor{SpringGreen}  $\tilde{p}_i < 0.01$ & 0.539 \\
\hline 
 \multirow{3}{*}{Cont}
	 & 0 & \cellcolor{SpringGreen}  $\tilde{p}_i < 0.01$ & 0.365 \\
	 & 1 & \cellcolor{SpringGreen}  $\tilde{p}_i < 0.01$ & 0.44 \\
	 & 42 & \cellcolor{SpringGreen}  $\tilde{p}_i < 0.01$ & 0.416 \\
\hline 
 \multirow{3}{*}{NotM}
	 & 0 & \cellcolor{SpringGreen}  $\tilde{p}_i < 0.01$ & 0.505 \\
	 & 1 & \cellcolor{SpringGreen}  $\tilde{p}_i < 0.01$ & 0.524 \\
	 & 42 & \cellcolor{SpringGreen}  $\tilde{p}_i < 0.01$ & 0.583 \\
\hline 
 \end{tabular} \caption{Maximum negative token log-probability}\end{subtable}\caption{Gemma-4 E2B Mann-Whitney U Test results and effect sizes for uncertainty metrics calculated with \textbf{the first token of the predicted label.}}\end{table*}

\begin{table*}[htbp] 
\subsection{Gemma-4 E4B}
\centering\begin{subtable}{0.48\textwidth}
\begin{tabular}{l|c||c|c}
\textbf{Label} & \textbf{Seed} & $\tilde{p}_i$ & \textbf{Effect size} \\
\hline
\hline

\multirow{3}{*}{Ent}
	 & 0 &  0.054 & 0.086 \\
	 & 1 & \cellcolor{SpringGreen}  0.017 & 0.102 \\
	 & 42 & \cellcolor{SpringGreen}  0.041 & 0.089 \\
\hline 
 \multirow{3}{*}{Cont}
	 & 0 & \cellcolor{SpringGreen}  $\tilde{p}_i < 0.01$ & 0.307 \\
	 & 1 & \cellcolor{SpringGreen}  $\tilde{p}_i < 0.01$ & 0.285 \\
	 & 42 & \cellcolor{SpringGreen}  $\tilde{p}_i < 0.01$ & 0.311 \\
\hline 
 \multirow{3}{*}{NotM}
	 & 0 &  1.0 & -0.443 \\
	 & 1 &  1.0 & -0.281 \\
	 & 42 &  1.0 & -0.29 \\
\hline 
 \end{tabular} \caption{Average negative log-probability}\end{subtable}\begin{subtable}{0.48\textwidth}
\begin{tabular}{l|c||c|c}
\textbf{Label} & \textbf{Seed} & $\tilde{p}_i$ & \textbf{Effect size} \\
\hline
\hline

\multirow{3}{*}{Ent}
	 & 0 & \cellcolor{SpringGreen}  $\tilde{p}_i < 0.01$ & 0.116 \\
	 & 1 & \cellcolor{SpringGreen}  $\tilde{p}_i < 0.01$ & 0.119 \\
	 & 42 & \cellcolor{SpringGreen}  $\tilde{p}_i < 0.01$ & 0.124 \\
\hline 
 \multirow{3}{*}{Cont}
	 & 0 & \cellcolor{SpringGreen}  $\tilde{p}_i < 0.01$ & 0.543 \\
	 & 1 & \cellcolor{SpringGreen}  $\tilde{p}_i < 0.01$ & 0.549 \\
	 & 42 & \cellcolor{SpringGreen}  $\tilde{p}_i < 0.01$ & 0.557 \\
\hline 
 \multirow{3}{*}{NotM}
	 & 0 &  1.0 & -0.01 \\
	 & 1 &  0.684 & 0.067 \\
	 & 42 &  0.965 & 0.007 \\
\hline 
 \end{tabular} \caption{Maximum negative token log-probability}\end{subtable}\caption{Gemma-4 E4B Mann-Whitney U Test results and effect sizes for uncertainty metrics calculated with \textbf{logits of the whole response}.}\end{table*}

 \begin{table*}[htbp] \centering\begin{subtable}{0.48\textwidth}
\begin{tabular}{l|c||c|c}
Label & Seed & $\tilde{p}_i$ & Effect size \\
\hline
\hline

\multirow{3}{*}{Ent}
	 & 0 & \cellcolor{SpringGreen}  $\tilde{p}_i < 0.01$ & 0.324 \\
	 & 1 & \cellcolor{SpringGreen}  $\tilde{p}_i < 0.01$ & 0.33 \\
	 & 42 & \cellcolor{SpringGreen}  $\tilde{p}_i < 0.01$ & 0.292 \\
\hline 
 \multirow{3}{*}{Cont}
	 & 0 & \cellcolor{SpringGreen}  $\tilde{p}_i < 0.01$ & 0.471 \\
	 & 1 & \cellcolor{SpringGreen}  $\tilde{p}_i < 0.01$ & 0.438 \\
	 & 42 & \cellcolor{SpringGreen}  $\tilde{p}_i < 0.01$ & 0.46 \\
\hline 
 \multirow{3}{*}{NotM}
	 & 0 & \cellcolor{SpringGreen}  $\tilde{p}_i < 0.01$ & 0.772 \\
	 & 1 & \cellcolor{SpringGreen}  $\tilde{p}_i < 0.01$ & 0.698 \\
	 & 42 & \cellcolor{SpringGreen}  $\tilde{p}_i < 0.01$ & 0.683 \\
\hline 
 \end{tabular} \caption{Average negative log-probability}\end{subtable}\begin{subtable}{0.48\textwidth}
\begin{tabular}{l|c||c|c}
Label & Seed & $\tilde{p}_i$ & Effect size \\
\hline
\hline

\multirow{3}{*}{Ent}
	 & 0 & \cellcolor{SpringGreen}  $\tilde{p}_i < 0.01$ & 0.314 \\
	 & 1 & \cellcolor{SpringGreen}  $\tilde{p}_i < 0.01$ & 0.325 \\
	 & 42 & \cellcolor{SpringGreen}  $\tilde{p}_i < 0.01$ & 0.286 \\
\hline 
 \multirow{3}{*}{Cont}
	 & 0 & \cellcolor{SpringGreen}  $\tilde{p}_i < 0.01$ & 0.473 \\
	 & 1 & \cellcolor{SpringGreen}  $\tilde{p}_i < 0.01$ & 0.438 \\
	 & 42 & \cellcolor{SpringGreen}  $\tilde{p}_i < 0.01$ & 0.457 \\
\hline 
 \multirow{3}{*}{NotM}
	 & 0 & \cellcolor{SpringGreen}  $\tilde{p}_i < 0.01$ & 0.772 \\
	 & 1 & \cellcolor{SpringGreen}  $\tilde{p}_i < 0.01$ & 0.698 \\
	 & 42 & \cellcolor{SpringGreen}  $\tilde{p}_i < 0.01$ & 0.684 \\
\hline 
 \end{tabular} \caption{Maximum negative token log-probability}\end{subtable}\caption{Gemma-4 E4B Mann-Whitney U Test results and effect sizes for uncertainty metrics calculated with \textbf{only the label logits}.}\end{table*}

\begin{table*}[htbp] \centering\begin{subtable}{0.48\textwidth}
\begin{tabular}{l|c||c|c}
Label & Seed & $\tilde{p}_i$ & Effect size \\
\hline
\hline

\multirow{3}{*}{Ent}
	 & 0 & \cellcolor{SpringGreen}  $\tilde{p}_i < 0.01$ & 0.325 \\
	 & 1 & \cellcolor{SpringGreen}  $\tilde{p}_i < 0.01$ & 0.332 \\
	 & 42 & \cellcolor{SpringGreen}  $\tilde{p}_i < 0.01$ & 0.294 \\
\hline 
 \multirow{3}{*}{Cont}
	 & 0 & \cellcolor{SpringGreen}  $\tilde{p}_i < 0.01$ & 0.443 \\
	 & 1 & \cellcolor{SpringGreen}  $\tilde{p}_i < 0.01$ & 0.406 \\
	 & 42 & \cellcolor{SpringGreen}  $\tilde{p}_i < 0.01$ & 0.424 \\
\hline 
 \multirow{3}{*}{NotM}
	 & 0 & \cellcolor{SpringGreen}  $\tilde{p}_i < 0.01$ & 0.772 \\
	 & 1 & \cellcolor{SpringGreen}  $\tilde{p}_i < 0.01$ & 0.698 \\
	 & 42 & \cellcolor{SpringGreen}  $\tilde{p}_i < 0.01$ & 0.684 \\
\hline 
 \end{tabular} \caption{Average negative log-probability}\end{subtable}\begin{subtable}{0.48\textwidth}
\begin{tabular}{l|c||c|c}
Label & Seed & $\tilde{p}_i$ & Effect size \\
\hline
\hline

\multirow{3}{*}{Ent}
	 & 0 & \cellcolor{SpringGreen}  $\tilde{p}_i < 0.01$ & 0.325 \\
	 & 1 & \cellcolor{SpringGreen}  $\tilde{p}_i < 0.01$ & 0.332 \\
	 & 42 & \cellcolor{SpringGreen}  $\tilde{p}_i < 0.01$ & 0.294 \\
\hline 
 \multirow{3}{*}{Cont}
	 & 0 & \cellcolor{SpringGreen}  $\tilde{p}_i < 0.01$ & 0.443 \\
	 & 1 & \cellcolor{SpringGreen}  $\tilde{p}_i < 0.01$ & 0.406 \\
	 & 42 & \cellcolor{SpringGreen}  $\tilde{p}_i < 0.01$ & 0.424 \\
\hline 
 \multirow{3}{*}{NotM}
	 & 0 & \cellcolor{SpringGreen}  $\tilde{p}_i < 0.01$ & 0.772 \\
	 & 1 & \cellcolor{SpringGreen}  $\tilde{p}_i < 0.01$ & 0.698 \\
	 & 42 & \cellcolor{SpringGreen}  $\tilde{p}_i < 0.01$ & 0.684 \\
\hline 
 \end{tabular} \caption{Maximum negative token log-probability}\end{subtable}\caption{Gemma-4 E4B Mann-Whitney U Test results and effect sizes for uncertainty metrics calculated with \textbf{the first token of the predicted label.}}\end{table*}


\begin{table*}[htbp] 
\subsection{Gemma-4 26B}
\centering\begin{subtable}{0.48\textwidth}
\begin{tabular}{l|c||c|c}
\textbf{Label} & \textbf{Seed} & $\tilde{p}_i$ & \textbf{Effect size} \\
\hline
\hline

\multirow{3}{*}{Ent}
	 & 0 & \cellcolor{SpringGreen}  $\tilde{p}_i < 0.01$ & 0.572 \\
	 & 1 & \cellcolor{SpringGreen}  $\tilde{p}_i < 0.01$ & 0.518 \\
	 & 42 & \cellcolor{SpringGreen}  $\tilde{p}_i < 0.01$ & 0.647 \\
\hline 
 \multirow{3}{*}{Cont}
	 & 0 &  0.062 & 0.51 \\
	 & 1 &  0.132 & 0.434 \\
	 & 42 &  0.407 & 0.265 \\
\hline 
 \multirow{3}{*}{NotM}
	 & 0 &  0.718 & 0.242 \\
	 & 1 &  0.562 & 0.421 \\
	 & 42 &  0.102 & 0.602 \\
\hline 
 \end{tabular} \caption{Average negative log-probability}\end{subtable}\begin{subtable}{0.48\textwidth}
\begin{tabular}{l|c||c|c}
\textbf{Label} & \textbf{Seed} & $\tilde{p}_i$ & \textbf{Effect size} \\
\hline
\hline

\multirow{3}{*}{Ent}
	 & 0 & \cellcolor{SpringGreen}  $\tilde{p}_i < 0.01$ & 0.46 \\
	 & 1 & \cellcolor{SpringGreen}  0.035 & 0.324 \\
	 & 42 & \cellcolor{SpringGreen}  $\tilde{p}_i < 0.01$ & 0.515 \\
\hline 
 \multirow{3}{*}{Cont}
	 & 0 &  0.5 & 0.235 \\
	 & 1 &  0.07 & 0.505 \\
	 & 42 &  0.746 & 0.128 \\
\hline 
 \multirow{3}{*}{NotM}
	 & 0 &  0.293 & 0.516 \\
	 & 1 &  0.746 & 0.25 \\
	 & 42 &  0.116 & 0.585 \\
\hline 
 \end{tabular} \caption{Maximum negative token log-probability}\end{subtable}\caption{Gemma-4 26B Mann-Whitney U Test results and effect sizes for uncertainty metrics calculated with \textbf{logits of the whole response}.}\end{table*}

 \begin{table*}[htbp] \centering\begin{subtable}{0.48\textwidth}
\begin{tabular}{l|c||c|c}
Label & Seed & $\tilde{p}_i$ & Effect size \\
\hline
\hline

\multirow{3}{*}{Ent}
	 & 0 &  0.899 & -0.151 \\
	 & 1 &  0.574 & 0.001 \\
	 & 42 &  0.509 & 0.023 \\
\hline 
 \multirow{3}{*}{Cont}
	 & 0 &  0.688 & -0.069 \\
	 & 1 & \cellcolor{SpringGreen}  0.038 & 0.465 \\
	 & 42 &  0.058 & 0.402 \\
\hline 
 \multirow{3}{*}{NotM}
	 & 0 & \cellcolor{SpringGreen}  0.01 & 0.826 \\
	 & 1 &  0.436 & 0.171 \\
	 & 42 &  0.244 & 0.295 \\
\hline 
 \end{tabular} \caption{Average negative log-probability}\end{subtable}\begin{subtable}{0.48\textwidth}
\begin{tabular}{l|c||c|c}
Label & Seed & $\tilde{p}_i$ & Effect size \\
\hline
\hline

\multirow{3}{*}{Ent}
	 & 0 &  0.969 & -0.234 \\
	 & 1 &  0.62 & -0.015 \\
	 & 42 &  0.795 & -0.085 \\
\hline 
 \multirow{3}{*}{Cont}
	 & 0 &  0.583 & 0.0 \\
	 & 1 & \cellcolor{SpringGreen}  0.019 & 0.535 \\
	 & 42 & \cellcolor{SpringGreen}  0.049 & 0.419 \\
\hline 
 \multirow{3}{*}{NotM}
	 & 0 & \cellcolor{SpringGreen}  0.014 & 0.808 \\
	 & 1 &  0.436 & 0.171 \\
	 & 42 &  0.274 & 0.267 \\
\hline 
 \end{tabular} \caption{Maximum negative token log-probability}\end{subtable}\caption{Gemma-4 26B Mann-Whitney U Test results and effect sizes for uncertainty metrics calculated with \textbf{only the label logits}.}\end{table*}

\begin{table*}[htbp] \centering\begin{subtable}{0.48\textwidth}
\begin{tabular}{l|c||c|c}
Label & Seed & $\tilde{p}_i$ & Effect size \\
\hline
\hline

\multirow{3}{*}{Ent}
	 & 0 &  0.538 & 0.01 \\
	 & 1 &  0.241 & 0.119 \\
	 & 42 &  0.136 & 0.164 \\
\hline 
 \multirow{3}{*}{Cont}
	 & 0 &  0.732 & -0.118 \\
	 & 1 &  0.335 & 0.152 \\
	 & 42 &  0.487 & 0.051 \\
\hline 
 \multirow{3}{*}{NotM}
	 & 0 & \cellcolor{SpringGreen}  0.011 & 0.817 \\
	 & 1 &  0.408 & 0.197 \\
	 & 42 &  0.242 & 0.284 \\
\hline 
 \end{tabular} \caption{Average negative log-probability}\end{subtable}\begin{subtable}{0.48\textwidth}
\begin{tabular}{l|c||c|c}
Label & Seed & $\tilde{p}_i$ & Effect size \\
\hline
\hline

\multirow{3}{*}{Ent}
	 & 0 &  0.538 & 0.01 \\
	 & 1 &  0.241 & 0.119 \\
	 & 42 &  0.136 & 0.164 \\
\hline 
 \multirow{3}{*}{Cont}
	 & 0 &  0.732 & -0.118 \\
	 & 1 &  0.335 & 0.152 \\
	 & 42 &  0.487 & 0.051 \\
\hline 
 \multirow{3}{*}{NotM}
	 & 0 & \cellcolor{SpringGreen}  0.011 & 0.817 \\
	 & 1 &  0.408 & 0.197 \\
	 & 42 &  0.242 & 0.284 \\
\hline 
 \end{tabular} \caption{Maximum negative token log-probability}\end{subtable}\caption{Gemma-4 26B Mann-Whitney U Test results and effect sizes for uncertainty metrics calculated with \textbf{the first token of the predicted label.}}\end{table*}


\begin{table*}[htbp] 
\subsubsection{Ministral-3 3B}
\centering\begin{subtable}{0.48\textwidth}
\begin{tabular}{l|c||c|c}
\textbf{Label} & \textbf{Seed} & $\tilde{p}_i$ & \textbf{Effect size} \\
\hline
\hline

\multirow{3}{*}{Ent}
	 & 0 &  1.0 & -0.163 \\
	 & 1 &  1.0 & -0.062 \\
	 & 42 &  1.0 & -0.107 \\
\hline 
 \multirow{3}{*}{Cont}
	 & 0 &  1.0 & 0.0 \\
	 & 1 &  0.746 & 0.667 \\
	 & 42 &  1.0 & 0.0 \\
\hline 
 \multirow{3}{*}{NotM}
	 & 0 &  1.0 & -0.029 \\
	 & 1 &  1.0 & -0.077 \\
	 & 42 &  0.992 & 0.006 \\
\hline 
 \end{tabular} \caption{Average negative log-probability}\end{subtable}\begin{subtable}{0.48\textwidth}
\begin{tabular}{l|c||c|c}
\textbf{Label} & \textbf{Seed} & $\tilde{p}_i$ & \textbf{Effect size} \\
\hline
\hline

\multirow{3}{*}{Ent}
	 & 0 &  0.445 & 0.105 \\
	 & 1 &  0.353 & 0.122 \\
	 & 42 &  1.0 & -0.101 \\
\hline 
 \multirow{3}{*}{Cont}
	 & 0 &  1.0 & 0.0 \\
	 & 1 &  0.746 & 0.667 \\
	 & 42 &  1.0 & -0.222 \\
\hline 
 \multirow{3}{*}{NotM}
	 & 0 &  0.731 & 0.125 \\
	 & 1 &  1.0 & -0.003 \\
	 & 42 &  1.0 & -0.035 \\
\hline 
 \end{tabular} \caption{Maximum negative token log-probability}\end{subtable}\caption{Ministral-3 3B Mann-Whitney U Test results and effect sizes for uncertainty metrics calculated with \textbf{logits of the whole response}.}\end{table*}

 \begin{table*}[htbp] \centering\begin{subtable}{0.48\textwidth}
\begin{tabular}{l|c||c|c}
Label & Seed & $\tilde{p}_i$ & Effect size \\
\hline
\hline

\multirow{3}{*}{Ent}
	 & 0 &  0.746 & -0.05 \\
	 & 1 &  0.574 & 0.001 \\
	 & 42 &  0.899 & -0.116 \\
\hline 
 \multirow{3}{*}{Cont}
	 & 0 &  0.37 & 0.4 \\
	 & 1 &  0.509 & 0.333 \\
	 & 42 &  0.688 & -0.111 \\
\hline 
 \multirow{3}{*}{NotM}
	 & 0 &  0.25 & 0.183 \\
	 & 1 &  0.583 & -0.003 \\
	 & 42 &  0.508 & 0.04 \\
\hline 
 \end{tabular} \caption{Average negative log-probability}\end{subtable}\begin{subtable}{0.48\textwidth}
\begin{tabular}{l|c||c|c}
Label & Seed & $\tilde{p}_i$ & Effect size \\
\hline
\hline

\multirow{3}{*}{Ent}
	 & 0 &  0.728 & -0.045 \\
	 & 1 &  0.574 & 0.001 \\
	 & 42 &  0.894 & -0.107 \\
\hline 
 \multirow{3}{*}{Cont}
	 & 0 &  0.37 & 0.4 \\
	 & 1 &  0.509 & 0.333 \\
	 & 42 &  0.688 & -0.111 \\
\hline 
 \multirow{3}{*}{NotM}
	 & 0 &  0.2 & 0.216 \\
	 & 1 &  0.651 & -0.044 \\
	 & 42 &  0.407 & 0.089 \\
\hline 
 \end{tabular} \caption{Maximum negative token log-probability}\end{subtable}\caption{Ministral-3 3B Mann-Whitney U Test results and effect sizes for uncertainty metrics calculated with \textbf{only the label logits}.}\end{table*}

 \begin{table*}[htbp] \centering\begin{subtable}{0.48\textwidth}
\begin{tabular}{l|c||c|c}
Label & Seed & $\tilde{p}_i$ & Effect size \\
\hline
\hline

\multirow{3}{*}{Ent}
	 & 0 &  0.715 & -0.045 \\
	 & 1 &  0.551 & 0.002 \\
	 & 42 &  0.876 & -0.105 \\
\hline 
 \multirow{3}{*}{Cont}
	 & 0 &  0.368 & 0.4 \\
	 & 1 &  0.5 & 0.333 \\
	 & 42 &  0.677 & -0.111 \\
\hline 
 \multirow{3}{*}{NotM}
	 & 0 &  0.201 & 0.212 \\
	 & 1 &  0.57 & -0.003 \\
	 & 42 &  0.43 & 0.073 \\
\hline 
 \end{tabular} \caption{Average negative log-probability}\end{subtable}\begin{subtable}{0.48\textwidth}
\begin{tabular}{l|c||c|c}
Label & Seed & $\tilde{p}_i$ & Effect size \\
\hline
\hline

\multirow{3}{*}{Ent}
	 & 0 &  0.715 & -0.045 \\
	 & 1 &  0.551 & 0.002 \\
	 & 42 &  0.876 & -0.105 \\
\hline 
 \multirow{3}{*}{Cont}
	 & 0 &  0.368 & 0.4 \\
	 & 1 &  0.5 & 0.333 \\
	 & 42 &  0.677 & -0.111 \\
\hline 
 \multirow{3}{*}{NotM}
	 & 0 &  0.201 & 0.212 \\
	 & 1 &  0.57 & -0.003 \\
	 & 42 &  0.43 & 0.073 \\
\hline 
 \end{tabular} \caption{Maximum negative token log-probability}\end{subtable}\caption{Ministral-3 3B Mann-Whitney U Test results and effect sizes for uncertainty metrics calculated with \textbf{the first token of the predicted label.}}\end{table*}


\begin{table*}[htbp]
\subsection{Ministral-3 8B}
\centering\begin{subtable}{0.48\textwidth}
\begin{tabular}{l|c||c|c}
\textbf{Label} & \textbf{Seed} & $\tilde{p}_i$ & \textbf{Effect size} \\
\hline
\hline

\multirow{3}{*}{Ent}
	 & 0 & \cellcolor{SpringGreen}  $\tilde{p}_i < 0.01$ & 0.418 \\
	 & 1 &  0.083 & 0.233 \\
	 & 42 &  0.054 & 0.261 \\
\hline 
 \multirow{3}{*}{Cont}
	 & 0 &  1.0 & -0.086 \\
	 & 1 &  1.0 & -0.035 \\
	 & 42 &  1.0 & -0.119 \\
\hline 
 \multirow{3}{*}{NotM}
	 & 0 &  0.834 & 0.096 \\
	 & 1 &  1.0 & -0.837 \\
	 & 42 &  1.0 & -0.437 \\
\hline 
 \end{tabular} \caption{Average negative log-probability}\end{subtable}\begin{subtable}{0.48\textwidth}
\begin{tabular}{l|c||c|c}
\textbf{Label} & \textbf{Seed} & $\tilde{p}_i$ & \textbf{Effect size} \\
\hline
\hline

\multirow{3}{*}{Ent}
	 & 0 &  1.0 & -0.163 \\
	 & 1 &  1.0 & -0.13 \\
	 & 42 &  1.0 & -0.133 \\
\hline 
 \multirow{3}{*}{Cont}
	 & 0 &  0.752 & 0.105 \\
	 & 1 &  0.888 & 0.06 \\
	 & 42 &  1.0 & -0.04 \\
\hline 
 \multirow{3}{*}{NotM}
	 & 0 & \cellcolor{SpringGreen}  $\tilde{p}_i < 0.01$ & 0.841 \\
	 & 1 & \cellcolor{SpringGreen}  0.029 & 0.652 \\
	 & 42 & \cellcolor{SpringGreen}  $\tilde{p}_i < 0.01$ & 0.749 \\
\hline 
 \end{tabular} \caption{Maximum negative token log-probability}\end{subtable}\caption{Ministral-3 8B Mann-Whitney U Test results and effect sizes for uncertainty metrics calculated with \textbf{logits of the whole response}.}\end{table*}

 \begin{table*}[htbp] \centering\begin{subtable}{0.48\textwidth}
\begin{tabular}{l|c||c|c}
Label & Seed & $\tilde{p}_i$ & Effect size \\
\hline
\hline

\multirow{3}{*}{Ent}
	 & 0 &  0.187 & 0.126 \\
	 & 1 &  0.207 & 0.117 \\
	 & 42 &  0.204 & 0.119 \\
\hline 
 \multirow{3}{*}{Cont}
	 & 0 &  0.288 & 0.162 \\
	 & 1 &  0.545 & 0.022 \\
	 & 42 &  0.275 & 0.183 \\
\hline 
 \multirow{3}{*}{NotM}
	 & 0 & \cellcolor{SpringGreen}  0.033 & 0.508 \\
	 & 1 &  0.372 & 0.148 \\
	 & 42 & \cellcolor{SpringGreen}  0.018 & 0.508 \\
\hline 
 \end{tabular} \caption{Average negative log-probability}\end{subtable}\begin{subtable}{0.48\textwidth}
\begin{tabular}{l|c||c|c}
Label & Seed & $\tilde{p}_i$ & Effect size \\
\hline
\hline

\multirow{3}{*}{Ent}
	 & 0 &  0.091 & 0.173 \\
	 & 1 &  0.204 & 0.119 \\
	 & 42 &  0.141 & 0.149 \\
\hline 
 \multirow{3}{*}{Cont}
	 & 0 &  0.299 & 0.156 \\
	 & 1 &  0.62 & -0.022 \\
	 & 42 &  0.285 & 0.175 \\
\hline 
 \multirow{3}{*}{NotM}
	 & 0 & \cellcolor{SpringGreen}  0.036 & 0.495 \\
	 & 1 &  0.353 & 0.163 \\
	 & 42 & \cellcolor{SpringGreen}  0.019 & 0.504 \\
\hline 
 \end{tabular} \caption{Maximum negative token log-probability}\end{subtable}\caption{Ministral-3 8B Mann-Whitney U Test results and effect sizes for uncertainty metrics calculated with \textbf{only the label logits}.}\end{table*}

 \begin{table*}[htbp] \centering\begin{subtable}{0.48\textwidth}
\begin{tabular}{l|c||c|c}
Label & Seed & $\tilde{p}_i$ & Effect size \\
\hline
\hline

\multirow{3}{*}{Ent}
	 & 0 &  0.13 & 0.151 \\
	 & 1 &  0.227 & 0.108 \\
	 & 42 &  0.14 & 0.146 \\
\hline 
 \multirow{3}{*}{Cont}
	 & 0 &  0.487 & 0.048 \\
	 & 1 &  0.606 & -0.022 \\
	 & 42 &  0.402 & 0.103 \\
\hline 
 \multirow{3}{*}{NotM}
	 & 0 & \cellcolor{SpringGreen}  $\tilde{p}_i < 0.01$ & 0.907 \\
	 & 1 & \cellcolor{SpringGreen}  $\tilde{p}_i < 0.01$ & 0.741 \\
	 & 42 & \cellcolor{SpringGreen}  $\tilde{p}_i < 0.01$ & 0.754 \\
\hline 
 \end{tabular} \caption{Average negative log-probability}\end{subtable}\begin{subtable}{0.48\textwidth}
\begin{tabular}{l|c||c|c}
Label & Seed & $\tilde{p}_i$ & Effect size \\
\hline
\hline

\multirow{3}{*}{Ent}
	 & 0 &  0.13 & 0.151 \\
	 & 1 &  0.227 & 0.108 \\
	 & 42 &  0.14 & 0.146 \\
\hline 
 \multirow{3}{*}{Cont}
	 & 0 &  0.487 & 0.048 \\
	 & 1 &  0.606 & -0.022 \\
	 & 42 &  0.402 & 0.103 \\
\hline 
 \multirow{3}{*}{NotM}
	 & 0 & \cellcolor{SpringGreen}  $\tilde{p}_i < 0.01$ & 0.907 \\
	 & 1 & \cellcolor{SpringGreen}  $\tilde{p}_i < 0.01$ & 0.741 \\
	 & 42 & \cellcolor{SpringGreen}  $\tilde{p}_i < 0.01$ & 0.754 \\
\hline 
 \end{tabular} \caption{Maximum negative token log-probability}\end{subtable}\caption{Ministral-3 8B Mann-Whitney U Test results and effect sizes for uncertainty metrics calculated with \textbf{the first token of the predicted label.}}\end{table*}


\begin{table*}[htbp]
\subsection{Ministral-3 14B}
\centering\begin{subtable}{0.48\textwidth}
\begin{tabular}{l|c||c|c}
\textbf{Label} & \textbf{Seed} & $\tilde{p}_i$ & \textbf{Effect size} \\
\hline
\hline

\multirow{3}{*}{Ent}
	 & 0 &  0.054 & 0.256 \\
	 & 1 &  0.337 & 0.153 \\
	 & 42 &  0.116 & 0.211 \\
\hline 
 \multirow{3}{*}{Cont}
	 & 0 &  0.377 & 0.292 \\
	 & 1 &  1.0 & -0.23 \\
	 & 42 &  0.532 & 0.208 \\
\hline 
 \multirow{3}{*}{NotM}
	 & 0 &  0.407 & 0.56 \\
	 & 1 &  1.0 & -0.698 \\
	 & 42 &  1.0 & -0.886 \\
\hline 
 \end{tabular} \caption{Average negative log-probability}\end{subtable}\begin{subtable}{0.48\textwidth}
\begin{tabular}{l|c||c|c}
\textbf{Label} & \textbf{Seed} & $\tilde{p}_i$ & \textbf{Effect size} \\
\hline
\hline

\multirow{3}{*}{Ent}
	 & 0 &  1.0 & -0.123 \\
	 & 1 &  1.0 & -0.137 \\
	 & 42 &  1.0 & -0.178 \\
\hline 
 \multirow{3}{*}{Cont}
	 & 0 &  1.0 & -0.303 \\
	 & 1 &  1.0 & -0.412 \\
	 & 42 &  1.0 & -0.285 \\
\hline 
 \multirow{3}{*}{NotM}
	 & 0 & \cellcolor{SpringGreen}  0.024 & 0.96 \\
	 & 1 & \cellcolor{SpringGreen}  0.029 & 0.953 \\
	 & 42 & \cellcolor{SpringGreen}  $\tilde{p}_i < 0.01$ & 1.0 \\
\hline 
 \end{tabular} \caption{Maximum negative token log-probability}\end{subtable}\caption{Ministral-3 14B Mann-Whitney U Test results and effect sizes for uncertainty metrics calculated with \textbf{logits of the whole response}.}\end{table*}

\begin{table*}[htbp] \centering\begin{subtable}{0.48\textwidth}
\begin{tabular}{l|c||c|c}
Label & Seed & $\tilde{p}_i$ & Effect size \\
\hline
\hline

\multirow{3}{*}{Ent}
	 & 0 & \cellcolor{SpringGreen}  $\tilde{p}_i < 0.01$ & 0.362 \\
	 & 1 &  0.282 & 0.092 \\
	 & 42 & \cellcolor{SpringGreen}  0.019 & 0.252 \\
\hline 
 \multirow{3}{*}{Cont}
	 & 0 &  0.091 & 0.364 \\
	 & 1 &  0.728 & -0.088 \\
	 & 42 &  0.171 & 0.262 \\
\hline 
 \multirow{3}{*}{NotM}
	 & 0 & \cellcolor{SpringGreen}  0.019 & 0.9 \\
	 & 1 & \cellcolor{SpringGreen}  0.038 & 0.837 \\
	 & 42 & \cellcolor{SpringGreen}  0.01 & 0.955 \\
\hline 
 \end{tabular} \caption{Average negative log-probability}\end{subtable}\begin{subtable}{0.48\textwidth}
\begin{tabular}{l|c||c|c}
Label & Seed & $\tilde{p}_i$ & Effect size \\
\hline
\hline

\multirow{3}{*}{Ent}
	 & 0 & \cellcolor{SpringGreen}  $\tilde{p}_i < 0.01$ & 0.356 \\
	 & 1 &  0.301 & 0.085 \\
	 & 42 & \cellcolor{SpringGreen}  0.023 & 0.245 \\
\hline 
 \multirow{3}{*}{Cont}
	 & 0 &  0.073 & 0.395 \\
	 & 1 &  0.67 & -0.049 \\
	 & 42 &  0.073 & 0.369 \\
\hline 
 \multirow{3}{*}{NotM}
	 & 0 & \cellcolor{SpringGreen}  0.019 & 0.9 \\
	 & 1 & \cellcolor{SpringGreen}  0.038 & 0.837 \\
	 & 42 & \cellcolor{SpringGreen}  0.01 & 0.955 \\
\hline 
 \end{tabular} \caption{Maximum negative token log-probability}\end{subtable}\caption{Ministral-3 14B Mann-Whitney U Test results and effect sizes for uncertainty metrics calculated with \textbf{only the label logits}.}\end{table*}

 \begin{table*}[htbp] \centering\begin{subtable}{0.48\textwidth}
\begin{tabular}{l|c||c|c}
Label & Seed & $\tilde{p}_i$ & Effect size \\
\hline
\hline

\multirow{3}{*}{Ent}
	 & 0 & \cellcolor{SpringGreen}  $\tilde{p}_i < 0.01$ & 0.356 \\
	 & 1 &  0.297 & 0.085 \\
	 & 42 & \cellcolor{SpringGreen}  0.022 & 0.245 \\
\hline 
 \multirow{3}{*}{Cont}
	 & 0 &  0.072 & 0.395 \\
	 & 1 &  0.655 & -0.049 \\
	 & 42 &  0.072 & 0.369 \\
\hline 
 \multirow{3}{*}{NotM}
	 & 0 & \cellcolor{SpringGreen}  0.019 & 0.9 \\
	 & 1 & \cellcolor{SpringGreen}  0.038 & 0.837 \\
	 & 42 & \cellcolor{SpringGreen}  $\tilde{p}_i < 0.01$ & 0.955 \\
\hline 
 \end{tabular} \caption{Average negative log-probability}\end{subtable}\begin{subtable}{0.48\textwidth}
\begin{tabular}{l|c||c|c}
Label & Seed & $\tilde{p}_i$ & Effect size \\
\hline
\hline

\multirow{3}{*}{Ent}
	 & 0 & \cellcolor{SpringGreen}  $\tilde{p}_i < 0.01$ & 0.356 \\
	 & 1 &  0.297 & 0.085 \\
	 & 42 & \cellcolor{SpringGreen}  0.022 & 0.245 \\
\hline 
 \multirow{3}{*}{Cont}
	 & 0 &  0.072 & 0.395 \\
	 & 1 &  0.655 & -0.049 \\
	 & 42 &  0.072 & 0.369 \\
\hline 
 \multirow{3}{*}{NotM}
	 & 0 & \cellcolor{SpringGreen}  0.019 & 0.9 \\
	 & 1 & \cellcolor{SpringGreen}  0.038 & 0.837 \\
	 & 42 & \cellcolor{SpringGreen}  $\tilde{p}_i < 0.01$ & 0.955 \\
\hline 
 \end{tabular} \caption{Maximum negative token log-probability}\end{subtable}\caption{Ministral-3 14B Mann-Whitney U Test results and effect sizes for uncertainty metrics calculated with \textbf{the first token of the predicted label.}}\end{table*}


\begin{table*}[htbp]
\subsection{Qwen-3.5 4B}
\centering\begin{subtable}{0.48\textwidth}
\begin{tabular}{l|c||c|c}
\textbf{Label} & \textbf{Seed} & $\tilde{p}_i$ & \textbf{Effect size} \\
\hline
\hline

\multirow{3}{*}{Ent}
	 & 0 &  1.0 & -0.057 \\
	 & 1 &  0.917 & 0.027 \\
	 & 42 &  0.684 & 0.087 \\
\hline 
 \multirow{3}{*}{Cont}
	 & 0 &  0.335 & 0.284 \\
	 & 1 &  1.0 & -0.041 \\
	 & 42 &  0.597 & 0.195 \\
\hline 
 \multirow{3}{*}{NotM}
	 & 0 &  1.0 & -0.136 \\
	 & 1 &  1.0 & -0.04 \\
	 & 42 &  1.0 & -0.516 \\
\hline 
 \end{tabular} \caption{Average negative log-probability}\end{subtable}\begin{subtable}{0.48\textwidth}
\begin{tabular}{l|c||c|c}
\textbf{Label} & \textbf{Seed} & $\tilde{p}_i$ & \textbf{Effect size} \\
\hline
\hline

\multirow{3}{*}{Ent}
	 & 0 &  1.0 & -0.087 \\
	 & 1 &  1.0 & -0.006 \\
	 & 42 &  0.746 & 0.066 \\
\hline 
 \multirow{3}{*}{Cont}
	 & 0 &  0.732 & 0.13 \\
	 & 1 &  0.944 & 0.027 \\
	 & 42 &  0.659 & 0.176 \\
\hline 
 \multirow{3}{*}{NotM}
	 & 0 &  1.0 & -0.073 \\
	 & 1 &  0.944 & 0.045 \\
	 & 42 &  1.0 & -0.144 \\
\hline 
 \end{tabular} \caption{Maximum negative token log-probability}\end{subtable}\caption{Qwen-3.5 4B Mann-Whitney U Test results and effect sizes for uncertainty metrics calculated with \textbf{logits of the whole response}.}\end{table*}

 \begin{table*}[htbp] \centering\begin{subtable}{0.48\textwidth}
\begin{tabular}{l|c||c|c}
Label & Seed & $\tilde{p}_i$ & Effect size \\
\hline
\hline

\multirow{3}{*}{Ent}
	 & 0 & \cellcolor{SpringGreen}  $\tilde{p}_i < 0.01$ & 0.395 \\
	 & 1 & \cellcolor{SpringGreen}  $\tilde{p}_i < 0.01$ & 0.42 \\
	 & 42 & \cellcolor{SpringGreen}  $\tilde{p}_i < 0.01$ & 0.341 \\
\hline 
 \multirow{3}{*}{Cont}
	 & 0 &  0.085 & 0.34 \\
	 & 1 &  0.728 & -0.095 \\
	 & 42 &  0.424 & 0.095 \\
\hline 
 \multirow{3}{*}{NotM}
	 & 0 & \cellcolor{SpringGreen}  0.019 & 0.691 \\
	 & 1 & \cellcolor{SpringGreen}  $\tilde{p}_i < 0.01$ & 0.802 \\
	 & 42 & \cellcolor{SpringGreen}  0.014 & 0.656 \\
\hline 
 \end{tabular} \caption{Average negative log-probability}\end{subtable}\begin{subtable}{0.48\textwidth}
\begin{tabular}{l|c||c|c}
Label & Seed & $\tilde{p}_i$ & Effect size \\
\hline
\hline

\multirow{3}{*}{Ent}
	 & 0 & \cellcolor{SpringGreen}  $\tilde{p}_i < 0.01$ & 0.408 \\
	 & 1 & \cellcolor{SpringGreen}  $\tilde{p}_i < 0.01$ & 0.419 \\
	 & 42 & \cellcolor{SpringGreen}  $\tilde{p}_i < 0.01$ & 0.344 \\
\hline 
 \multirow{3}{*}{Cont}
	 & 0 &  0.08 & 0.347 \\
	 & 1 &  0.713 & -0.082 \\
	 & 42 &  0.407 & 0.104 \\
\hline 
 \multirow{3}{*}{NotM}
	 & 0 & \cellcolor{SpringGreen}  0.018 & 0.7 \\
	 & 1 & \cellcolor{SpringGreen}  $\tilde{p}_i < 0.01$ & 0.808 \\
	 & 42 & \cellcolor{SpringGreen}  0.014 & 0.656 \\
\hline 
 \end{tabular} \caption{Maximum negative token log-probability}\end{subtable}\caption{Qwen-3.5 4B Mann-Whitney U Test results and effect sizes for uncertainty metrics calculated with \textbf{only the label logits}.}\end{table*}

 \begin{table*}[htbp] \centering\begin{subtable}{0.48\textwidth}
\begin{tabular}{l|c||c|c}
Label & Seed & $\tilde{p}_i$ & Effect size \\
\hline
\hline

\multirow{3}{*}{Ent}
	 & 0 & \cellcolor{SpringGreen}  $\tilde{p}_i < 0.01$ & 0.408 \\
	 & 1 & \cellcolor{SpringGreen}  $\tilde{p}_i < 0.01$ & 0.419 \\
	 & 42 & \cellcolor{SpringGreen}  $\tilde{p}_i < 0.01$ & 0.344 \\
\hline 
 \multirow{3}{*}{Cont}
	 & 0 &  0.078 & 0.347 \\
	 & 1 &  0.698 & -0.082 \\
	 & 42 &  0.405 & 0.104 \\
\hline 
 \multirow{3}{*}{NotM}
	 & 0 & \cellcolor{SpringGreen}  0.017 & 0.7 \\
	 & 1 & \cellcolor{SpringGreen}  $\tilde{p}_i < 0.01$ & 0.808 \\
	 & 42 & \cellcolor{SpringGreen}  0.013 & 0.656 \\
\hline 
 \end{tabular} \caption{Average negative log-probability}\end{subtable}\begin{subtable}{0.48\textwidth}
\begin{tabular}{l|c||c|c}
Label & Seed & $\tilde{p}_i$ & Effect size \\
\hline
\hline

\multirow{3}{*}{Ent}
	 & 0 & \cellcolor{SpringGreen}  $\tilde{p}_i < 0.01$ & 0.408 \\
	 & 1 & \cellcolor{SpringGreen}  $\tilde{p}_i < 0.01$ & 0.419 \\
	 & 42 & \cellcolor{SpringGreen}  $\tilde{p}_i < 0.01$ & 0.344 \\
\hline 
 \multirow{3}{*}{Cont}
	 & 0 &  0.078 & 0.347 \\
	 & 1 &  0.698 & -0.082 \\
	 & 42 &  0.405 & 0.104 \\
\hline 
 \multirow{3}{*}{NotM}
	 & 0 & \cellcolor{SpringGreen}  0.017 & 0.7 \\
	 & 1 & \cellcolor{SpringGreen}  $\tilde{p}_i < 0.01$ & 0.808 \\
	 & 42 & \cellcolor{SpringGreen}  0.013 & 0.656 \\
\hline 
 \end{tabular} \caption{Maximum negative token log-probability}\end{subtable}\caption{Qwen-3.5 4B Mann-Whitney U Test results and effect sizes for uncertainty metrics calculated with \textbf{the first token of the predicted label.}}\end{table*}


\begin{table*}[htbp]
\subsection{Qwen-3.5 9B}
\centering\begin{subtable}{0.48\textwidth}
\begin{tabular}{l|c||c|c}
\textbf{Label} & \textbf{Seed} & $\tilde{p}_i$ & \textbf{Effect size} \\
\hline
\hline

\multirow{3}{*}{Ent}
	 & 0 &  0.062 & 0.307 \\
	 & 1 &  0.175 & 0.225 \\
	 & 42 &  0.211 & 0.229 \\
\hline 
 \multirow{3}{*}{Cont}
	 & 0 &  0.07 & 0.511 \\
	 & 1 &  0.752 & 0.125 \\
	 & 42 & \cellcolor{SpringGreen}  0.024 & 0.636 \\
\hline 
 \multirow{3}{*}{NotM}
	 & 0 &  1.0 & -0.236 \\
	 & 1 &  1.0 & -0.428 \\
	 & 42 &  1.0 & -0.249 \\
\hline 
 \end{tabular} \caption{Average negative log-probability}\end{subtable}\begin{subtable}{0.48\textwidth}
\begin{tabular}{l|c||c|c}
\textbf{Label} & \textbf{Seed} & $\tilde{p}_i$ & \textbf{Effect size} \\
\hline
\hline

\multirow{3}{*}{Ent}
	 & 0 &  1.0 & -0.265 \\
	 & 1 &  1.0 & -0.298 \\
	 & 42 &  1.0 & -0.204 \\
\hline 
 \multirow{3}{*}{Cont}
	 & 0 &  1.0 & -0.078 \\
	 & 1 &  0.888 & 0.062 \\
	 & 42 &  1.0 & -0.057 \\
\hline 
 \multirow{3}{*}{NotM}
	 & 0 &  1.0 & -0.576 \\
	 & 1 &  1.0 & -0.542 \\
	 & 42 &  1.0 & -0.57 \\
\hline 
 \end{tabular} \caption{Maximum negative token log-probability}\end{subtable}\caption{Qwen-3.5 9B Mann-Whitney U Test results and effect sizes for uncertainty metrics calculated with \textbf{logits of the whole response}.}\end{table*}

 \begin{table*}[htbp] \centering\begin{subtable}{0.48\textwidth}
\begin{tabular}{l|c||c|c}
Label & Seed & $\tilde{p}_i$ & Effect size \\
\hline
\hline

\multirow{3}{*}{Ent}
	 & 0 & \cellcolor{SpringGreen}  0.036 & 0.284 \\
	 & 1 & \cellcolor{SpringGreen}  0.014 & 0.324 \\
	 & 42 &  0.13 & 0.197 \\
\hline 
 \multirow{3}{*}{Cont}
	 & 0 &  0.407 & 0.111 \\
	 & 1 &  0.35 & 0.146 \\
	 & 42 &  0.5 & 0.057 \\
\hline 
 \multirow{3}{*}{NotM}
	 & 0 & \cellcolor{SpringGreen}  $\tilde{p}_i < 0.01$ & 0.75 \\
	 & 1 & \cellcolor{SpringGreen}  $\tilde{p}_i < 0.01$ & 0.599 \\
	 & 42 & \cellcolor{SpringGreen}  $\tilde{p}_i < 0.01$ & 0.658 \\
\hline 
 \end{tabular} \caption{Average negative log-probability}\end{subtable}\begin{subtable}{0.48\textwidth}
\begin{tabular}{l|c||c|c}
Label & Seed & $\tilde{p}_i$ & Effect size \\
\hline
\hline

\multirow{3}{*}{Ent}
	 & 0 & \cellcolor{SpringGreen}  0.033 & 0.29 \\
	 & 1 & \cellcolor{SpringGreen}  0.014 & 0.324 \\
	 & 42 &  0.13 & 0.197 \\
\hline 
 \multirow{3}{*}{Cont}
	 & 0 &  0.126 & 0.333 \\
	 & 1 &  0.332 & 0.156 \\
	 & 42 &  0.299 & 0.182 \\
\hline 
 \multirow{3}{*}{NotM}
	 & 0 & \cellcolor{SpringGreen}  $\tilde{p}_i < 0.01$ & 0.75 \\
	 & 1 & \cellcolor{SpringGreen}  $\tilde{p}_i < 0.01$ & 0.589 \\
	 & 42 & \cellcolor{SpringGreen}  $\tilde{p}_i < 0.01$ & 0.655 \\
\hline 
 \end{tabular} \caption{Maximum negative token log-probability}\end{subtable}\caption{Qwen-3.5 9B Mann-Whitney U Test results and effect sizes for uncertainty metrics calculated with \textbf{only the label logits}.}\end{table*}

 \begin{table*}[htbp] \centering\begin{subtable}{0.48\textwidth}
\begin{tabular}{l|c||c|c}
Label & Seed & $\tilde{p}_i$ & Effect size \\
\hline
\hline

\multirow{3}{*}{Ent}
	 & 0 & \cellcolor{SpringGreen}  0.032 & 0.29 \\
	 & 1 & \cellcolor{SpringGreen}  0.013 & 0.324 \\
	 & 42 &  0.13 & 0.197 \\
\hline 
 \multirow{3}{*}{Cont}
	 & 0 &  0.069 & 0.411 \\
	 & 1 &  0.225 & 0.229 \\
	 & 42 &  0.207 & 0.25 \\
\hline 
 \multirow{3}{*}{NotM}
	 & 0 & \cellcolor{SpringGreen}  $\tilde{p}_i < 0.01$ & 0.75 \\
	 & 1 & \cellcolor{SpringGreen}  $\tilde{p}_i < 0.01$ & 0.589 \\
	 & 42 & \cellcolor{SpringGreen}  $\tilde{p}_i < 0.01$ & 0.655 \\
\hline 
 \end{tabular} \caption{Average negative log-probability}\end{subtable}\begin{subtable}{0.48\textwidth}
\begin{tabular}{l|c||c|c}
Label & Seed & $\tilde{p}_i$ & Effect size \\
\hline
\hline

\multirow{3}{*}{Ent}
	 & 0 & \cellcolor{SpringGreen}  0.032 & 0.29 \\
	 & 1 & \cellcolor{SpringGreen}  0.013 & 0.324 \\
	 & 42 &  0.13 & 0.197 \\
\hline 
 \multirow{3}{*}{Cont}
	 & 0 &  0.069 & 0.411 \\
	 & 1 &  0.225 & 0.229 \\
	 & 42 &  0.207 & 0.25 \\
\hline 
 \multirow{3}{*}{NotM}
	 & 0 & \cellcolor{SpringGreen}  $\tilde{p}_i < 0.01$ & 0.75 \\
	 & 1 & \cellcolor{SpringGreen}  $\tilde{p}_i < 0.01$ & 0.589 \\
	 & 42 & \cellcolor{SpringGreen}  $\tilde{p}_i < 0.01$ & 0.655 \\
\hline 
 \end{tabular} \caption{Maximum negative token log-probability}\end{subtable}\caption{Qwen-3.5 9B Mann-Whitney U Test results and effect sizes for uncertainty metrics calculated with \textbf{the first token of the predicted label.}}\end{table*}


\begin{table*}[htbp]
\subsection{Qwen-3.5 27B}
\centering\begin{subtable}{0.48\textwidth}
\begin{tabular}{l|c||c|c}
\textbf{Label} & \textbf{Seed} & $\tilde{p}_i$ & \textbf{Effect size} \\
\hline
\hline

\multirow{3}{*}{Ent}
	 & 0 &  0.692 & 0.032 \\
	 & 1 &  0.185 & 0.081 \\
	 & 42 & \cellcolor{SpringGreen}  $\tilde{p}_i < 0.01$ & 0.158 \\
\hline 
 \multirow{3}{*}{Cont}
	 & 0 &  1.0 & -0.035 \\
	 & 1 &  1.0 & -0.06 \\
	 & 42 &  1.0 & -0.009 \\
\hline 
 \multirow{3}{*}{NotM}
	 & 0 &  1.0 & -0.111 \\
	 & 1 &  1.0 & -0.264 \\
	 & 42 &  1.0 & -0.304 \\
\hline 
 \end{tabular} \caption{Average negative log-probability}\end{subtable}\begin{subtable}{0.48\textwidth}
\begin{tabular}{l|c||c|c}
\textbf{Label} & \textbf{Seed} & $\tilde{p}_i$ & \textbf{Effect size} \\
\hline
\hline

\multirow{3}{*}{Ent}
	 & 0 &  1.0 & -0.235 \\
	 & 1 &  1.0 & -0.153 \\
	 & 42 &  1.0 & -0.182 \\
\hline 
 \multirow{3}{*}{Cont}
	 & 0 &  1.0 & -0.32 \\
	 & 1 &  1.0 & -0.284 \\
	 & 42 &  1.0 & -0.268 \\
\hline 
 \multirow{3}{*}{NotM}
	 & 0 &  1.0 & -0.128 \\
	 & 1 &  1.0 & -0.042 \\
	 & 42 &  1.0 & -0.041 \\
\hline 
 \end{tabular} \caption{Maximum negative token log-probability}\end{subtable}\caption{Qwen-3.5 27B Mann-Whitney U Test results and effect sizes for uncertainty metrics calculated with \textbf{logits of the whole response}.}\end{table*}

\begin{table*}[htbp] \centering\begin{subtable}{0.48\textwidth}
\begin{tabular}{l|c||c|c}
Label & Seed & $\tilde{p}_i$ & Effect size \\
\hline
\hline

\multirow{3}{*}{Ent}
	 & 0 & \cellcolor{SpringGreen}  $\tilde{p}_i < 0.01$ & 0.371 \\
	 & 1 & \cellcolor{SpringGreen}  $\tilde{p}_i < 0.01$ & 0.417 \\
	 & 42 & \cellcolor{SpringGreen}  $\tilde{p}_i < 0.01$ & 0.448 \\
\hline 
 \multirow{3}{*}{Cont}
	 & 0 &  0.483 & 0.017 \\
	 & 1 & \cellcolor{SpringGreen}  $\tilde{p}_i < 0.01$ & 0.183 \\
	 & 42 & \cellcolor{SpringGreen}  $\tilde{p}_i < 0.01$ & 0.227 \\
\hline 
 \multirow{3}{*}{NotM}
	 & 0 & \cellcolor{SpringGreen}  $\tilde{p}_i < 0.01$ & 0.487 \\
	 & 1 & \cellcolor{SpringGreen}  $\tilde{p}_i < 0.01$ & 0.293 \\
	 & 42 & \cellcolor{SpringGreen}  $\tilde{p}_i < 0.01$ & 0.367 \\
\hline 
 \end{tabular} \caption{Average negative log-probability}\end{subtable}\begin{subtable}{0.48\textwidth}
\begin{tabular}{l|c||c|c}
Label & Seed & $\tilde{p}_i$ & Effect size \\
\hline
\hline

\multirow{3}{*}{Ent}
	 & 0 & \cellcolor{SpringGreen}  $\tilde{p}_i < 0.01$ & 0.376 \\
	 & 1 & \cellcolor{SpringGreen}  $\tilde{p}_i < 0.01$ & 0.416 \\
	 & 42 & \cellcolor{SpringGreen}  $\tilde{p}_i < 0.01$ & 0.447 \\
\hline 
 \multirow{3}{*}{Cont}
	 & 0 &  0.438 & 0.023 \\
	 & 1 & \cellcolor{SpringGreen}  $\tilde{p}_i < 0.01$ & 0.182 \\
	 & 42 & \cellcolor{SpringGreen}  $\tilde{p}_i < 0.01$ & 0.238 \\
\hline 
 \multirow{3}{*}{NotM}
	 & 0 & \cellcolor{SpringGreen}  $\tilde{p}_i < 0.01$ & 0.549 \\
	 & 1 & \cellcolor{SpringGreen}  $\tilde{p}_i < 0.01$ & 0.388 \\
	 & 42 & \cellcolor{SpringGreen}  $\tilde{p}_i < 0.01$ & 0.434 \\
\hline 
 \end{tabular} \caption{Maximum negative token log-probability}\end{subtable}\caption{Qwen-3.5 27B Mann-Whitney U Test results and effect sizes for uncertainty metrics calculated with \textbf{only the label logits}.}\end{table*}

 \begin{table*}[htbp] \centering\begin{subtable}{0.48\textwidth}
\begin{tabular}{l|c||c|c}
Label & Seed & $\tilde{p}_i$ & Effect size \\
\hline
\hline

\multirow{3}{*}{Ent}
	 & 0 & \cellcolor{SpringGreen}  $\tilde{p}_i < 0.01$ & 0.376 \\
	 & 1 & \cellcolor{SpringGreen}  $\tilde{p}_i < 0.01$ & 0.416 \\
	 & 42 & \cellcolor{SpringGreen}  $\tilde{p}_i < 0.01$ & 0.447 \\
\hline 
 \multirow{3}{*}{Cont}
	 & 0 &  0.43 & 0.023 \\
	 & 1 & \cellcolor{SpringGreen}  $\tilde{p}_i < 0.01$ & 0.182 \\
	 & 42 & \cellcolor{SpringGreen}  $\tilde{p}_i < 0.01$ & 0.242 \\
\hline 
 \multirow{3}{*}{NotM}
	 & 0 & \cellcolor{SpringGreen}  $\tilde{p}_i < 0.01$ & 0.55 \\
	 & 1 & \cellcolor{SpringGreen}  $\tilde{p}_i < 0.01$ & 0.387 \\
	 & 42 & \cellcolor{SpringGreen}  $\tilde{p}_i < 0.01$ & 0.435 \\
\hline 
 \end{tabular} \caption{Average negative log-probability}\end{subtable}\begin{subtable}{0.48\textwidth}
\begin{tabular}{l|c||c|c}
Label & Seed & $\tilde{p}_i$ & Effect size \\
\hline
\hline

\multirow{3}{*}{Ent}
	 & 0 & \cellcolor{SpringGreen}  $\tilde{p}_i < 0.01$ & 0.376 \\
	 & 1 & \cellcolor{SpringGreen}  $\tilde{p}_i < 0.01$ & 0.416 \\
	 & 42 & \cellcolor{SpringGreen}  $\tilde{p}_i < 0.01$ & 0.447 \\
\hline 
 \multirow{3}{*}{Cont}
	 & 0 &  0.43 & 0.023 \\
	 & 1 & \cellcolor{SpringGreen}  $\tilde{p}_i < 0.01$ & 0.182 \\
	 & 42 & \cellcolor{SpringGreen}  $\tilde{p}_i < 0.01$ & 0.242 \\
\hline 
 \multirow{3}{*}{NotM}
	 & 0 & \cellcolor{SpringGreen}  $\tilde{p}_i < 0.01$ & 0.55 \\
	 & 1 & \cellcolor{SpringGreen}  $\tilde{p}_i < 0.01$ & 0.387 \\
	 & 42 & \cellcolor{SpringGreen}  $\tilde{p}_i < 0.01$ & 0.435 \\
\hline 
 \end{tabular} \caption{Maximum negative token log-probability}\end{subtable}\caption{Qwen-3.5 27B Mann-Whitney U Test results and effect sizes for uncertainty metrics calculated with \textbf{the first token of the predicted label.}}\end{table*}


\begin{table*}[htbp]
\subsection{Qwen-3.6 35B}
\centering\begin{subtable}{0.48\textwidth}
\begin{tabular}{l|c||c|c}
\textbf{Label} & \textbf{Seed} & $\tilde{p}_i$ & \textbf{Effect size} \\
\hline
\hline

\multirow{3}{*}{Ent}
	 & 0 &  1.0 & -0.071 \\
	 & 1 &  1.0 & -0.268 \\
	 & 42 &  1.0 & -0.243 \\
\hline 
 \multirow{3}{*}{Cont}
	 & 0 &  1.0 & -0.037 \\
	 & 1 &  1.0 & -0.165 \\
	 & 42 &  1.0 & -0.378 \\
\hline 
 \multirow{3}{*}{NotM}
	 & 0 &  0.944 & 0.056 \\
	 & 1 &  1.0 & -0.13 \\
	 & 42 &  0.888 & 0.099 \\
\hline 
 \end{tabular} \caption{Average negative log-probability}\end{subtable}\begin{subtable}{0.48\textwidth}
\begin{tabular}{l|c||c|c}
\textbf{Label} & \textbf{Seed} & $\tilde{p}_i$ & \textbf{Effect size} \\
\hline
\hline

\multirow{3}{*}{Ent}
	 & 0 &  0.138 & 0.205 \\
	 & 1 &  0.176 & 0.199 \\
	 & 42 & \cellcolor{SpringGreen}  0.046 & 0.233 \\
\hline 
 \multirow{3}{*}{Cont}
	 & 0 &  0.888 & 0.042 \\
	 & 1 &  0.834 & 0.059 \\
	 & 42 &  0.661 & 0.11 \\
\hline 
 \multirow{3}{*}{NotM}
	 & 0 &  0.843 & 0.157 \\
	 & 1 &  0.834 & 0.174 \\
	 & 42 &  1.0 & -0.149 \\
\hline 
 \end{tabular} \caption{Maximum negative token log-probability}\end{subtable}\caption{Qwen-3.6 35B Mann-Whitney U Test results and effect sizes for uncertainty metrics calculated with \textbf{logits of the whole response}.}\end{table*}

 \begin{table*}[htbp] \centering\begin{subtable}{0.48\textwidth}
\begin{tabular}{l|c||c|c}
Label & Seed & $\tilde{p}_i$ & Effect size \\
\hline
\hline

\multirow{3}{*}{Ent}
	 & 0 &  0.106 & 0.167 \\
	 & 1 &  0.826 & -0.096 \\
	 & 42 &  0.193 & 0.108 \\
\hline 
 \multirow{3}{*}{Cont}
	 & 0 &  0.277 & 0.129 \\
	 & 1 &  0.136 & 0.214 \\
	 & 42 &  0.053 & 0.266 \\
\hline 
 \multirow{3}{*}{NotM}
	 & 0 &  0.648 & -0.079 \\
	 & 1 &  0.918 & -0.587 \\
	 & 42 &  0.187 & 0.411 \\
\hline 
 \end{tabular} \caption{Average negative log-probability}\end{subtable}\begin{subtable}{0.48\textwidth}
\begin{tabular}{l|c||c|c}
Label & Seed & $\tilde{p}_i$ & Effect size \\
\hline
\hline

\multirow{3}{*}{Ent}
	 & 0 &  0.074 & 0.189 \\
	 & 1 &  0.802 & -0.085 \\
	 & 42 &  0.145 & 0.126 \\
\hline 
 \multirow{3}{*}{Cont}
	 & 0 &  0.187 & 0.178 \\
	 & 1 &  0.089 & 0.252 \\
	 & 42 & \cellcolor{SpringGreen}  0.036 & 0.293 \\
\hline 
 \multirow{3}{*}{NotM}
	 & 0 &  0.554 & 0.034 \\
	 & 1 &  0.899 & -0.533 \\
	 & 42 &  0.145 & 0.468 \\
\hline 
 \end{tabular} \caption{Maximum negative token log-probability}\end{subtable}\caption{Qwen-3.6 35B Mann-Whitney U Test results and effect sizes for uncertainty metrics calculated with \textbf{only the label logits}.}\end{table*}

\begin{table*}[htbp] \centering\begin{subtable}{0.48\textwidth}
\begin{tabular}{l|c||c|c}
Label & Seed & $\tilde{p}_i$ & Effect size \\
\hline
\hline

\multirow{3}{*}{Ent}
	 & 0 &  0.072 & 0.189 \\
	 & 1 &  0.79 & -0.085 \\
	 & 42 &  0.14 & 0.126 \\
\hline 
 \multirow{3}{*}{Cont}
	 & 0 &  0.18 & 0.178 \\
	 & 1 &  0.087 & 0.252 \\
	 & 42 & \cellcolor{SpringGreen}  0.035 & 0.293 \\
\hline 
 \multirow{3}{*}{NotM}
	 & 0 &  0.538 & 0.034 \\
	 & 1 &  0.889 & -0.533 \\
	 & 42 &  0.14 & 0.468 \\
\hline 
 \end{tabular} \caption{Average negative log-probability}\end{subtable}\begin{subtable}{0.48\textwidth}
\begin{tabular}{l|c||c|c}
Label & Seed & $\tilde{p}_i$ & Effect size \\
\hline
\hline

\multirow{3}{*}{Ent}
	 & 0 &  0.072 & 0.189 \\
	 & 1 &  0.79 & -0.085 \\
	 & 42 &  0.14 & 0.126 \\
\hline 
 \multirow{3}{*}{Cont}
	 & 0 &  0.18 & 0.178 \\
	 & 1 &  0.087 & 0.252 \\
	 & 42 & \cellcolor{SpringGreen}  0.035 & 0.293 \\
\hline 
 \multirow{3}{*}{NotM}
	 & 0 &  0.538 & 0.034 \\
	 & 1 &  0.889 & -0.533 \\
	 & 42 &  0.14 & 0.468 \\
\hline 
 \end{tabular} \caption{Maximum negative token log-probability}\end{subtable}\caption{Qwen-3.6 35B Mann-Whitney U Test results and effect sizes for uncertainty metrics calculated with \textbf{the first token of the predicted label.}}\end{table*}

\clearpage
\begin{table*}[h]
\section{NLI4Wills results} \label{app:nli4wills-results}

\subsection{NLI4Wills-Idaho}
\centering
\begin{tabular}{l||cccc|c}
\textbf{Model} & \textbf{Precision} & \textbf{Recall} & \textbf{F1} & \textbf{Accuracy} & \textbf{Invalid Rate}\\
\hline
\hline
Majority   & 0.197 & 0.333 & 0.248 & 0.591 & ---
\\
\hline
Gemma-4 E2B $^{\dag}$
    & $0.766 \pm 0.011$
    & $0.722 \pm 0.023$
    & $0.713 \pm 0.025$
    & $0.786 \pm 0.016$
    & $0.0 \pm 0.0$ \\ 
Gemma-4 E4B $^{\dag}$
    & $0.808 \pm 0.011$
    & $0.751 \pm 0.010$
    & $0.768 \pm 0.015$
    & $0.840 \pm 0.006$
    & $0.0 \pm 0.0$ \\ 
Gemma-4 26B $^{\ast}$
    & $0.591 \pm 0.014$
    & $0.590 \pm 0.001$
    & $0.584 \pm 0.007$
    & $0.766 \pm 0.005$
    & $0.1 \pm 0.0$ \\ 
\hline
Ministral-3 3B $^{\dag}$
    & $0.618 \pm 0.028$
    & $0.598 \pm 0.017$
    & $0.516 \pm 0.026$
    & $0.526 \pm 0.028$
    & $0.0 \pm 0.0$ \\ 
Ministral-3 8B $^{\dag}$
    & $0.795 \pm 0.006$
    & $0.806 \pm 0.009$
    & $0.800 \pm 0.008$
    & $0.848 \pm 0.006$
    & $0.0 \pm 0.0$ \\ 
Ministral-3 14B $^{\dag}$
    & $0.804 \pm 0.031$
    & $0.789 \pm 0.026$
    & $0.785 \pm 0.029$
    & $0.842 \pm 0.021$
    & $0.0 \pm 0.0$ \\
\hline
Qwen-3.5 4B $^{\dag}$
    & $0.775 \pm 0.071$
    & $0.765 \pm 0.058$
    & $0.768 \pm 0.063$
    & $0.883 \pm 0.028$
    & $0.0 \pm 0.0$ \\ 
Qwen-3.5 9B $^{\dag}$
    & $0.861 \pm 0.018$
    & $0.824 \pm 0.025$
    & $0.830 \pm 0.024$
    & $0.874 \pm 0.017$
    & $0.0 \pm 0.0$ \\ 
Qwen-3.5 27B$^{\dag\ast}$
    & $0.763 \pm 0.110$
    & $0.699 \pm 0.109$
    & $0.722 \pm 0.111$
    & $0.900 \pm 0.012$
    & $0.0 \pm 0.0$ \\ 
\hline
\rowcolor{SpringGreen}
Qwen-3.6 35B$^{\dag\ast}$
    & $\mathbf{0.884 \pm 0.005}$
    & $\mathbf{0.864 \pm 0.009}$
    & $\mathbf{0.873 \pm 0.006}$
    & $\mathbf{0.903 \pm 0.000}$
    & $0.0 \pm 0.0$ \\ 
  \end{tabular}\\
  { $^\ast$ Reasoning enabled \hspace{20mm}
  $^\dag$ Output structure applied}
  \caption{NLI4Wills Idaho results. Each metric reports the average and standard deviation across three seeds. Boldface indicates the best-performing model per metric, and the ``Invalid Rate (\%)'' column indicates the average percentage of answers outputted by the LLM from which we were unable to retrieve a coherent label. The green row highlights the best-performing LLM. Majority baseline results are given for the \textit{unrelated} label.}
  \label{tab:nli4wills-idaho-results}

\end{table*}

\begin{table*}
\subsection{NLI4Wills-Tennessee}
\centering
\begin{tabular}{l||cccc|c}
\textbf{Model} & \textbf{Precision} & \textbf{Recall} & \textbf{F1} & \textbf{Accuracy} & \textbf{Invalid Rate}\\
\hline
Majority   & 0.182 & 0.333 & 0.235 & 0.545 & --- \\
\hline
RoBERTa-large-mnli & \textbf{0.967} & \textbf{0.962} & \textbf{0.964} & \textbf{0.969} & --- \\
\hline
Gemma-4 E2B
    & $0.676 \pm 0.011$
    & $0.616 \pm 0.005$
    & $0.616 \pm 0.004$
    & $0.701 \pm 0.008$
    & $0.0 \pm 0.0$ \\ 
Gemma-4 E4B
    & $0.785 \pm 0.021$
    & $0.673 \pm 0.014$
    & $0.702 \pm 0.016$
    & $0.766 \pm 0.012$
    & $0.0 \pm 0.0$ \\ 
Gemma-4 26B$^{\ast\dag}$
    & $0.553 \pm 0.005$
    & $0.497 \pm 0.006$
    & $0.521 \pm 0.005$
    & $0.685 \pm 0.005$
    & $0.1 \pm 0.0$ \\ 
\hline
Ministral-3 3B
    & $0.560 \pm 0.031$
    & $0.562 \pm 0.035$
    & $0.503 \pm 0.031$
    & $0.499 \pm 0.035$
    & $0.0 \pm 0.0$ \\ 
Ministral-3 8B
    & $0.684 \pm 0.009$
    & $0.662 \pm 0.001$
    & $0.671 \pm 0.003$
    & $0.720 \pm 0.002$
    & $0.0 \pm 0.0$ \\ 
Ministral-3 14B
    & $0.770 \pm 0.006$
    & $0.703 \pm 0.009$
    & $0.728 \pm 0.008$
    & $0.769 \pm 0.003$
    & $0.0 \pm 0.0$ \\ 
\hline
Qwen-3.5 4B
    & $0.769 \pm 0.012$
    & $0.724 \pm 0.019$
    & $0.742 \pm 0.017$
    & $0.779 \pm 0.010$
    & $0.0 \pm 0.0$ \\ 
Qwen-3.5 9B
    & $0.789 \pm 0.094$
    & $0.719 \pm 0.088$
    & $0.745 \pm 0.090$
    & $0.841 \pm 0.005$
    & $0.0 \pm 0.0$ \\ 
Qwen-3.5 27B
    & $0.649 \pm 0.002$
    & $0.547 \pm 0.006$
    & $0.580 \pm 0.005$
    & $0.808 \pm 0.006$
    & $0.0 \pm 0.0$ $^{\ast}$\\ 
\hline
\rowcolor{SpringGreen}
Qwen-3.6 35B
    & $0.831 \pm 0.018$
    & $0.771 \pm 0.008$
    & $0.794 \pm 0.010$
    & $0.818 \pm 0.008$
    & $0.0 \pm 0.0$ $^{\ast}$\\ 
  \end{tabular}\\
  { $^\ast$ Reasoning enabled \hspace{20mm}
  $^\dag$ Output structure applied}
  \caption{NLI4Wills Tennessee results. Each metric reports the average and standard deviation across three seeds. Boldface indicates the best-performing model per metric, and the ``Invalid Rate (\%)'' column indicates the average percentage of answers outputted by the LLM from which we were unable to retrieve a coherent label. The green row highlights the best-performing LLM. Majority baseline results are given for the \textit{unrelated} label.}
  \label{tab:nli4wills-tennessee-results}
\end{table*}

\clearpage

\begin{table*}[h]
\section{Model-specific results}\label{app:model-specific-metrics}

This section contains confusion matrices and ablation studies per model. We did not perform an ablation study for Ministral-3 models, as they do not have a reasoning mode.

\subsection{Gemma-4}
\subsubsection*{Gemma-4 E2B}
    \centering
    \begin{tabular}{r||ccc||c}
         \textbf{Setting} & \textbf{Accuracy} & \textbf{F1(C)} & \textbf{F1(E)} & \textbf{Invalid Rate (\%)} \\
         \hline\hline
         Reasoning &
         0.680 ± 0.013 & \textbf{0.268 ± 0.026} & \textbf{0.692 ± 0.008}
         & 0.0 ± 0.0\\
         Structure$^{*}$ &
         \textbf{0.688 ± 0.003} & 0.254 ± 0.018 & 0.681 ± 0.001 & 0.0 ± 0.0 \\
         Both &
         0.688 ± 0.011 & 0.246 ± 0.016 & 0.683 ± 0.008
         & 0.0 ± 0.0 \\
    \end{tabular}
    \caption{Gemma-4 E2B ablation results on the ContractNLI test set. The asterisk ($^{*}$) indicates the setting used in the main results.}
    \label{tab:ablation-gemma4-e2b}
\end{table*}

 \begin{table*}[]
    \centering
    \begin{tabular}{c||c|c|c|c}
 \textbf{True \textbackslash{} Pred} & \textbf{NotMentioned} & \textbf{Entailment} & \textbf{Contradiction} & \textbf{InvalidAnswer}  \\
\hline\hline
 NotMentioned & 455 ± 3 & 364 ± 4 & 84 ± 4 & 0 ± 0  \\
 Entailment & 66 ± 5 & 861 ± 4 & 41 ± 2 & 0 ± 0  \\
 Contradiction & 44 ± 6 & 54 ± 5 & 122 ± 2 & 0 ± 0
    \end{tabular}
    \caption{Gemma-4 E2B confusion matrix for the main results (Table \ref{tab:abridged-.3}).}
    \label{tab:confusion-gemma4-e2b}
\end{table*}

\begin{table*}[h]
    \centering
\subsubsection*{Gemma-4 E4B}
    \begin{tabular}{r||ccc||c}
         \textbf{Setting} & \textbf{Accuracy} & \textbf{F1(C)} & \textbf{F1(E)} & \textbf{Invalid Rate (\%)} \\
         \hline\hline
         Reasoning &
         0.744 ± 0.004 & \textbf{0.386 ± 0.013} & 0.747 ± 0.004
         & 0.0 ± 0.0 \\
         Structure$^{*}$ &
         \textbf{0.746 ± 0.006} & 0.343 ± 0.012 & \textbf{0.753 ± 0.005} & 0.0 ± 0.0  \\
         Both &
         0.738 ± 0.000 & 0.357 ± 0.013 & 0.744 ± 0.003
         & 0.0 ± 0.0 \\
    \end{tabular}
    \caption{Gemma-4 E4B ablation results on the ContractNLI test set. The asterisk ($^{*}$) indicates the setting used in the main results.}
    \label{tab:ablation-gemma4-e4b}
\end{table*}

 \begin{table*}[]
    \centering
    \begin{tabular}{c||c|c|c|c}
 \textbf{True \textbackslash{} Pred} & \textbf{NotMentioned} & \textbf{Entailment} & \textbf{Contradiction} & \textbf{InvalidAnswer}  \\
\hline\hline
 NotMentioned & 513 ± 10 & 289 ± 6 & 101 ± 5 & 0 ± 0  \\
 Entailment & 32 ± 3 & 902 ± 4 & 34 ± 2 & 0 ± 0  \\
 Contradiction & 23 ± 1 & 52 ± 2 & 145 ± 1 & 0 ± 0  \\
    \end{tabular}
    \caption{Gemma-4 E4B confusion matrix for the main results (Table \ref{tab:abridged-.3}).}
    \label{tab:confusion-gemma4-e4b}
\end{table*}

\begin{table*}[h]
\subsubsection*{Gemma-4 26B}
    \centering
    \begin{tabular}{r||ccc||c}
         \textbf{Setting} & \textbf{Accuracy} & \textbf{F1(C)} & \textbf{F1(E)} & \textbf{Invalid Rate (\%)}   \\
         \hline\hline
         Reasoning$^{*}$ &
         0.812 ± 0.002 & 0.443 ± 0.030 & \textbf{0.808 ± 0.003} & 1.0 ± 0.1  \\
         Structure &
         \textbf{0.827 ± 0.003} & \textbf{0.456 ± 0.016} & 0.803 ± 0.001
         & 0.0 ± 0.0  \\
         %
    \end{tabular}
    \caption{Gemma-4 26B ablation results on the ContractNLI test set. The asterisk ($^{*}$) indicates the setting used in the main results. No results are presented for reasoning-enabled structured outputs, as this leads to a $\sim 100\%$ invalid output rate.}
    \label{tab:ablation-gemma4-26b}
\end{table*}

 \begin{table*}[]
    \centering
    \begin{tabular}{c||c|c|c|c}
 \textbf{True \textbackslash{} Pred} & \textbf{NotMentioned} & \textbf{Entailment} & \textbf{Contradiction} & \textbf{InvalidAnswer}  \\
\hline\hline
 NotMentioned & 654 ± 1 & 143 ± 1 & 95 ± 2 & 12 ± 1  \\
 Entailment & 35 ± 2 & 893 ± 3 & 33 ± 3 & 7 ± 2  \\
 Contradiction & 16 ± 1 & 51 ± 2 & 152 ± 1 & 2 ± 1  \\
    \end{tabular}
    \caption{Gemma-4 26B confusion matrix for the main results (Table \ref{tab:abridged-.3}).}
    \label{tab:confusion-gemma4-26b}
\end{table*}


 \begin{table*}[]
\subsection{Ministral-3}

\subsubsection*{Ministral-3 3B}
    \centering
    \begin{tabular}{c||c|c|c|c}
 \textbf{True \textbackslash{} Pred} & \textbf{NotMentioned} & \textbf{Entailment} & \textbf{Contradiction} & \textbf{InvalidAnswer}  \\
\hline\hline
 NotMentioned & 322 ± 4 & 535 ± 5 & 46 ± 3 & 0 ± 0  \\
 Entailment & 67 ± 2 & 847 ± 4 & 54 ± 2 & 0 ± 0  \\
 Contradiction & 37 ± 6 & 149 ± 7 & 33 ± 5 & 0 ± 0  \\
    \end{tabular}
    \caption{Ministral-3 3B confusion matrix for the main results (Table \ref{tab:abridged-.3}).}
    \label{tab:confusion-ministral3-3b}
\end{table*}


 \begin{table*}[]
 \subsubsection*{Ministral-3 8B}
    \centering
    \begin{tabular}{c||c|c|c|c}
 \textbf{True \textbackslash{} Pred} & \textbf{NotMentioned} & \textbf{Entailment} & \textbf{Contradiction} & \textbf{InvalidAnswer}  \\
\hline\hline
 NotMentioned & 458 ± 9 & 337 ± 9 & 108 ± 3 & 0 ± 0  \\
 Entailment & 74 ± 6 & 832 ± 6 & 62 ± 3 & 0 ± 0  \\
 Contradiction & 33 ± 1 & 50 ± 2 & 136 ± 3 & 0 ± 0  \\
    \end{tabular}
    \caption{Ministral-3 8B confusion matrix for the main results (Table \ref{tab:abridged-.3}).}
    \label{tab:confusion-ministral3-8b}
\end{table*}


 \begin{table*}[]
 \subsubsection*{Ministral-3 14B}
    \centering
    \begin{tabular}{c||c|c|c|c}
 \textbf{True \textbackslash{} Pred} & \textbf{NotMentioned} & \textbf{Entailment} & \textbf{Contradiction} & \textbf{InvalidAnswer}  \\
\hline\hline
 NotMentioned & 447 ± 3 & 293 ± 4 & 163 ± 2 & 0 ± 0  \\
 Entailment & 24 ± 3 & 888 ± 1 & 56 ± 3 & 0 ± 0  \\
 Contradiction & 17 ± 1 & 34 ± 2 & 169 ± 2 & 0 ± 0  \\
    \end{tabular}
    \caption{Ministral-3 14B confusion matrix for the main results (Table \ref{tab:abridged-.3}).}
    \label{tab:confusion-ministral3-14b}
\end{table*}

\begin{table*}[h]
\subsection{Qwen-3.5}
\subsubsection*{Qwen-3.5 4B}
    \centering
    \begin{tabular}{r||ccc||c}
         \textbf{Setting} & \textbf{Accuracy} & \textbf{F1(C)} & \textbf{F1(E)} & \textbf{Invalid Rate (\%)}   \\
         \hline\hline
         Reasoning &
         0.519 ± 0.013 & 0.325 ± 0.033 & 0.616 ± 0.006
         & 36.3 ± 1.7 \\
         Structure$^{*}$ &
         0.741 ± 0.007 & \textbf{0.370 ± 0.025} & \textbf{0.742 ± 0.006}
         & 0.1 ± 0.1 \\
         Both &
         \textbf{0.750 ± 0.004} & 0.354 ± 0.046 & 0.737 ± 0.009
         & 3.5 ± 0.5 \\
    \end{tabular}
    \caption{Qwen-3.5 4B ablation results on the ContractNLI test set. The asterisk ($^{*}$) indicates the setting used in the main results.}
    \label{tab:ablation-qwen3.5-4b}
\end{table*}

 \begin{table*}[]
    \centering
    \begin{tabular}{c||c|c|c|c}
 \textbf{True \textbackslash{} Pred} & \textbf{NotMentioned} & \textbf{Entailment} & \textbf{Contradiction} & \textbf{InvalidAnswer}  \\
\hline\hline
 NotMentioned & 551 ± 12 & 206 ± 16 & 144 ± 4 & 2 ± 1  \\
 Entailment & 57 ± 2 & 845 ± 6 & 67 ± 4 & 0 ± 0  \\
 Contradiction & 25 ± 2 & 41 ± 2 & 154 ± 3 & 0 ± 0  \\
    \end{tabular}
    \caption{Qwen-3.5 4B confusion matrix for the main results (Table \ref{tab:abridged-.3}).}
    \label{tab:confusion-qwen3.5-4b}
\end{table*}

\begin{table*}[h]
\subsubsection*{Qwen-3.5 9B}
    \centering
    \begin{tabular}{r||ccc||c}
         \textbf{Setting} & \textbf{Accuracy} & \textbf{F1(C)} & \textbf{F1(E)} & \textbf{Invalid Rate (\%)}  \\
         \hline\hline
         Reasoning &
         0.704 ± 0.008 & \textbf{0.340 ± 0.028} & 0.748 ± 0.005
         & 12.0 ± 0.6 \\
         Structure$^{*}$ &
         \textbf{0.759 ± 0.004} & 0.330 ± 0.002 & \textbf{0.749 ± 0.002} & 0.0 ± 0.0    \\
         Both &
         0.729 ± 0.020 & 0.339 ± 0.049 & 0.722 ± 0.022
         & 0.2 ± 0.1  \\
    \end{tabular}
    \caption{Qwen-3.5 9B ablation results on the ContractNLI test set. The asterisk ($^{*}$) indicates the setting used in the main results.}
    \label{tab:ablation-qwen3.5-9b}
\end{table*}

 \begin{table*}[]
    \centering
    \begin{tabular}{c||c|c|c|c}
 \textbf{True \textbackslash{} Pred} & \textbf{NotMentioned} & \textbf{Entailment} & \textbf{Contradiction} & \textbf{InvalidAnswer}  \\
\hline\hline
 NotMentioned & 606 ± 4 & 184 ± 4 & 113 ± 8 & 0 ± 0  \\
 Entailment & 84 ± 7 & 827 ± 7 & 57 ± 9 & 0 ± 0  \\
 Contradiction & 22 ± 0 & 45 ± 2 & 154 ± 2 & 0 ± 0  \\
    \end{tabular}
    \caption{Qwen-3.5 9B confusion matrix for the main results (Table \ref{tab:abridged-.3}).}
    \label{tab:confusion-qwen3.5-9b}
\end{table*}

\begin{table*}[h]
\subsubsection*{Qwen-3.5 27B}
    \centering
    \begin{tabular}{r||ccc||c}
         \textbf{Setting} & \textbf{Accuracy} & \textbf{F1(C)} & \textbf{F1(E)} & \textbf{Invalid Rate (\%)}   \\
         \hline\hline
         Reasoning &
         0.260 ± 0.002 & 0.178 ± 0.059 & 0.424 ± 0.002
         & 68.6 ± 0.0 \\
         Structure &
         \textbf{0.796 ± 0.003} & 0.401 ± 0.019 & \textbf{0.786 ± 0.003}
         & 0.0 ± 0.0  \\
         Both$^{*}$ &
         0.793 ± 0.005 & \textbf{0.422 ± 0.024} & 0.783 ± 0.006 & 0.3 ± 0.0    \\
    \end{tabular}
    \caption{Qwen-3.5 27B ablation results on the ContractNLI test set. The asterisk ($^{*}$) indicates the setting used in the main results.}
    \label{tab:ablation-qwen3.5-27b}
\end{table*}

 \begin{table*}[]
    \centering
    \begin{tabular}{c||c|c|c|c}
 \textbf{True \textbackslash{} Pred} & \textbf{NotMentioned} & \textbf{Entailment} & \textbf{Contradiction} & \textbf{InvalidAnswer}  \\
\hline\hline
 NotMentioned & 636 ± 7 & 144 ± 4 & 120 ± 4 & 3 ± 2  \\
 Entailment & 51 ± 5 & 858 ± 6 & 56 ± 3 & 3 ± 1  \\
 Contradiction & 11 ± 3 & 44 ± 0 & 164 ± 2 & 1 ± 0  \\
     \end{tabular}
    \caption{Qwen-3.5 27B confusion matrix for the main results (Table \ref{tab:abridged-.3}).}
    \label{tab:confusion-qwen3.5-27b}
\end{table*}

\begin{table*}[h]
\subsection{Qwen-3.6}
\subsubsection*{Qwen-3.6 35B}
    \centering
    \begin{tabular}{r||ccc||c}
         \textbf{Setting} & \textbf{Accuracy} & \textbf{F1(C)} & \textbf{F1(E)} & \textbf{Invalid Rate (\%)}   \\
         \hline\hline
         Reasoning &
         \textbf{0.791 ± 0.005} & \textbf{0.387 ± 0.021} & \textbf{0.793 ± 0.003}
         & 0.1 ± 0.1  \\
         Structure &
         0.774 ± 0.005 & 0.364 ± 0.020 & 0.776 ± 0.002
         & 0.1 ± 0.0  \\
         Both$^{*}$ &
         0.760 ± 0.006 & 0.330 ± 0.005 & 0.780 ± 0.008 & 0.6 ± 0.2     \\
    \end{tabular}
    \caption{Qwen-3.6 35B ablation results on the ContractNLI test set. The asterisk ($^{*}$) indicates the setting used in the main results.}
    \label{tab:ablation-qwen3.6-35b}
\end{table*}

 \begin{table*}[]
    \centering
    \begin{tabular}{c||c|c|c|c}
 \textbf{True \textbackslash{} Pred} & \textbf{NotMentioned} & \textbf{Entailment} & \textbf{Contradiction} & \textbf{InvalidAnswer}  \\
\hline\hline
 NotMentioned & 520 ± 7 & 197 ± 12 & 175 ± 4 & 10 ± 3  \\
 Entailment & 23 ± 3 & 897 ± 6 & 45 ± 2 & 3 ± 1  \\
 Contradiction & 3 ± 1 & 46 ± 1 & 171 ± 1 & 0 ± 0  \\
    \end{tabular}
    \caption{Qwen-3.6 35B confusion matrix for the main results (Table \ref{tab:abridged-.3}).}
    \label{tab:confusion-qwen3.6-35b}
\end{table*}

\clearpage

\begin{table*}[h]
\section{ContractNLI class-specific metrics}\label{app:class-specific-metrics}
    \centering
    \begin{tabular}{c||c|c|c}
    \textbf{Model} & \textbf{P(C)} & \textbf{R(C)} & \textbf{F1(C)}  \\
    \hline\hline
         Gemma-4 E2B  & 0.215 ± 0.027 & 0.254 ± 0.012 & 0.254 ± 0.018\\
Gemma-4 E4B           & 0.203 ± 0.009 & 0.382 ± 0.044 & 0.343 ± 0.012 \\
Gemma-4 26B           & \textbf{0.287 ± 0.013} & 0.472 ± 0.051 & \textbf{0.443 ± 0.030}\\
\hline
Ministral-3 3B   & 0.124 ± 0.026 & 0.128 ± 0.062 & 0.110 ± 0.019 \\
Ministral-3 8B   & 0.219 ± 0.040 & 0.427 ± 0.029 & 0.325 ± 0.045 \\
Ministral-3 14B  & 0.190 ± 0.006 & \textbf{0.558 ± 0.009} & 0.372 ± 0.008 \\
\hline
Qwen-3.5 4B           & 0.208 ± 0.021 & 0.407 ± 0.019 & 0.370 ± 0.025 \\
Qwen-3.5 9B           & 0.187 ± 0.002 & 0.377 ± 0.008 & 0.330 ± 0.002 \\
Qwen-3.5 27B          & 0.255 ± 0.020 & 0.534 ± 0.026 & 0.422 ± 0.024 \\
\hline
Qwen-3.6 35B          & 0.156 ± 0.003 & 0.534 ± 0.004 & 0.330 ± 0.005 \\
    \end{tabular}
    \caption{\textbf{Contradiction} class metrics on the main results (Table \ref{tab:abridged-.3}). Boldface indicates the best-performing model per metric. All scores are micro averaged over documents, then macro averaged over labels.}
    \label{tab:metrics-contradiction}
\end{table*}

\begin{table*}[h]
    \centering
    \begin{tabular}{c||c|c|c}
    \textbf{Model} & \textbf{P(E)} & \textbf{R(E)} & \textbf{F1(E)}\\
    \hline\hline
         Gemma-4 E2B &  0.601 ± 0.002 & 0.837 ± 0.006 & 0.681 ± 0.001 \\
Gemma-4 E4B &          0.708 ± 0.007 & \textbf{0.868 ± 0.005} & 0.753 ± 0.005  \\
Gemma-4 26B     &      \textbf{0.781 ± 0.005} & 0.867 ± 0.002 & 0.808 ± 0.003 \\
\hline
Ministral-3 3B   &  0.541 ± 0.001 & 0.866 ± 0.008 & 0.628 ± 0.001 \\
Ministral-3 8B  &   0.639 ± 0.011 & 0.830 ± 0.007 & \textbf{0.691 ± 0.009} \\
Ministral-3 14B &   0.672 ± 0.004 & 0.851 ± 0.001 & 0.728 ± 0.002 \\
\hline
Qwen-3.5 4B       &     0.730 ± 0.009 & 0.796 ± 0.006 & 0.742 ± 0.006 \\
Qwen-3.5 9B      &      0.757 ± 0.002 & 0.804 ± 0.003 & 0.749 ± 0.002 \\
Qwen-3.5 27B    &       \textbf{0.781 ± 0.003} & 0.821 ± 0.008 & 0.783 ± 0.006 \\

Qwen-3.6 35B    &     0.747 ± 0.006 & 0.860 ± 0.010 & 0.780 ± 0.008
    \end{tabular}
    \caption{\textbf{Entailment} class metrics on the main results (Table \ref{tab:abridged-.3}). Boldface indicates the best-performing model per metric. All scores are micro averaged over documents, then macro averaged over labels.}
    \label{tab:metrics-entailment}
\end{table*}

\begin{table*}[h]
    \centering
    \begin{tabular}{c||c|c|c}
    \textbf{Model} & \textbf{P(N)} & \textbf{R(N)} & \textbf{F1(N)} \\
    \hline\hline
         Gemma-4 E2B                                                & 0.779 ± 0.019 & 0.430 ± 0.004 & 0.511 ± 0.004 \\
Gemma-4 E4B                                                         & 0.844 ± 0.010 & 0.491 ± 0.009 & 0.575 ± 0.007 \\
Gemma-4 26B          & \textbf{0.887 ± 0.006} & \textbf{0.657 ± 0.004} & \textbf{0.718 ± 0.003} \\
\hline
Ministral-3 3B  &   0.649 ± 0.041 & 0.292 ± 0.005 & 0.369 ± 0.011 \\
Ministral-3 8B  &   0.741 ± 0.031 & 0.431 ± 0.012 & 0.496 ± 0.016 \\
Ministral-3 14B  &  0.726 ± 0.047 & 0.394 ± 0.007 & 0.467 ± 0.014 \\
\hline
Qwen-3.5 4B      &      0.772 ± 0.015 & 0.538 ± 0.017 & 0.597 ± 0.015 \\
Qwen-3.5 9B      &      0.788 ± 0.034 & 0.624 ± 0.007 & 0.657 ± 0.012 \\
Qwen-3.5 27B     &      0.821 ± 0.033 & 0.630 ± 0.016 & 0.676 ± 0.018 \\
\hline
Qwen-3.6 35B      &   0.880 ± 0.007 & 0.498 ± 0.011 & 0.582 ± 0.013
    \end{tabular}
    \caption{\textbf{NotMentioned} class metrics on the main results (Table \ref{tab:abridged-.3}). Boldface indicates the best-performing model per metric. All scores are micro averaged over documents, then macro averaged over labels.}
    \label{tab:metrics-notmentioned}
\end{table*}




\end{document}